\documentclass[ACS,STIX2COL]{WileyNJD-v2}

\articletype{Article Type}%

\received{26 April 2016}
\revised{6 June 2016}
\accepted{6 June 2016}

\usepackage{makecell}   % Enables \Xhline
\usepackage{booktabs}   % For nice table rules (e.g., \toprule, \midrule, \bottomrule)

\begin{document}

\title{Novel Extraction of Discriminative Fine-Grained Feature to Improve Retinal Vessel Segmentation}%\protect\thanks{This is an example for title footnote.}}

\author[1,2,3,4,7,*]{Shuang Zeng}
\author[1,*]{Chee Hong Lee}
\author[5]{Micky C Nnamdi}
\author[6]{Wenqi Shi}
\author[5]{J Ben Tamo}
\author[1,2,3,4]{Lei Zhu}
\author[1,2,3,4]{Hangzhou He}
\author[1,2,3,4]{Xinliang Zhang}
\author[1,2,3,4]{Qian Chen}
\author[7]{May D. Wang}
\author[1,2,3,4]{Yanye Lu*}
\author[1,2,3,4]{Qiushi Ren*}

\authormark{Shuang Zeng \textsc{et al}}

\address[1]{\orgdiv{Department of Biomedical Engineering}, \orgname{Peking University}, \orgaddress{\state{Beijing}, \country{China}}}

\address[2]{\orgdiv{Institute of Medical Technology}, \orgname{Peking University Health Science Center, Peking University}, \orgaddress{\state{Beijing}, \country{China}}}

\address[3]{\orgdiv{National Biomedical Imaging Center}, \orgname{Peking University}, \orgaddress{\state{Beijing}, \country{China}}}

\address[4]{\orgdiv{Institute of Biomedical Engineering}, \orgname{Peking University Shenzhen Graduate School}, \orgaddress{\state{Shenzhen}, \country{China}}}

\address[5]{\orgdiv{school of Electrical and Computer Engineering}, \orgname{Georgia Institute of Technology}, \orgaddress{\state{Atlanta, GA}, \country{USA}}}

\address[6]{\orgdiv{Peter O’Donnell Jr. School of Public Health}, \orgname{at UT Southwestern Medical Center (UTSW)}, \orgaddress{\state{Dallas, TX}, \country{USA}}}

\address[7]{\orgdiv{Wallace H. Coulter Department of Biomedical Engineering}, \orgname{Georgia Institute of Technology and Emory University}, \orgaddress{\state{Atlanta, GA}, \country{USA}}}

\address[*]{Shuang Zeng and Chee Hong Lee 
 contributed equally to this work.}

\corres{*Yanye Lu, Qiushi Ren. \email{yanye.lu@pku.edu.cn, qren@pku.edu.cn}}

% \presentaddress{This is sample for present address text this is sample for present address text}

\abstract{Retinal vessel segmentation is a vital early detection method for several severe ocular diseases, which often manifests through changes in retinal vascular morphology. Despite significant progress in retinal vessel segmentation with the advancement of Convolutional Neural Networks (CNNs), there are still challenges to overcome. Specifically, retinal vessel segmentation aims to predict the class label for every pixel within a fundus image, with a primary focus on intra-image discrimination, making it vital for models to extract as many discriminative features as possible. Nevertheless, existing methods primarily focus on minimizing the difference between the output from the decoder and the label, but ignore making full use of feature-level fine-grained representations from the encoder. To address these issues, we propose a novel Attention U-shaped Kolmogorov-Arnold Network named AttUKAN for retinal vessel segmentation. Specifically, we implement Attention Gates (AGs) into  Kolmogorov-Arnold Networks (KANs) to enhance model sensitivity by suppressing irrelevant feature activations and model interpretability by non-linear modeling of KAN blocks. Additionally, we also design a novel Label-guided Pixel-wise Contrastive Loss (LPCL) to supervise our proposed AttUKAN to extract more discriminative features by distinguishing between foreground vessel-pixel sample pairs and background sample pairs. Experiments are conducted across four public datasets including  DRIVE, STARE, CHASE\_DB1, HRF and our private dataset. AttUKAN achieves F1 scores of 82.50\%, 81.14\%, 81.34\%, 80.21\% and 80.09\%, along with MIoU scores of 70.24\%, 68.64\%, 68.59\%, 67.21\% and 66.94\% in the above datasets, which are the highest compared to 11 networks for retinal vessel segmentation. Quantitative and qualitative results show that our AttUKAN achieves state-of-the-art performance and outperforms existing retinal vessel segmentation methods. Our code will be available at https://github.com/stevezs315/AttUKAN.}


% \jnlcitation{\cname{%
% \author{Williams K.}, 
% \author{B. Hoskins}, 
% \author{R. Lee}, 
% \author{G. Masato}, and 
% \author{T. Woollings}} (\cyear{2016}), 
% \ctitle{A regime analysis of Atlantic winter jet variability applied to evaluate HadGEM3-GC2}, \cjournal{Q.J.R. Meteorol. Soc.}, \cvol{2017;00:1--6}.}

\maketitle

%\footnotetext{\textbf{Abbreviations:} ANA, anti-nuclear antibodies; APC, antigen-presenting cells; IRF, interferon regulatory factor}

\section{Introduction}
\label{sec:introduction}

\begin{figure}[t]
\centering
\includegraphics[width=0.49\textwidth]{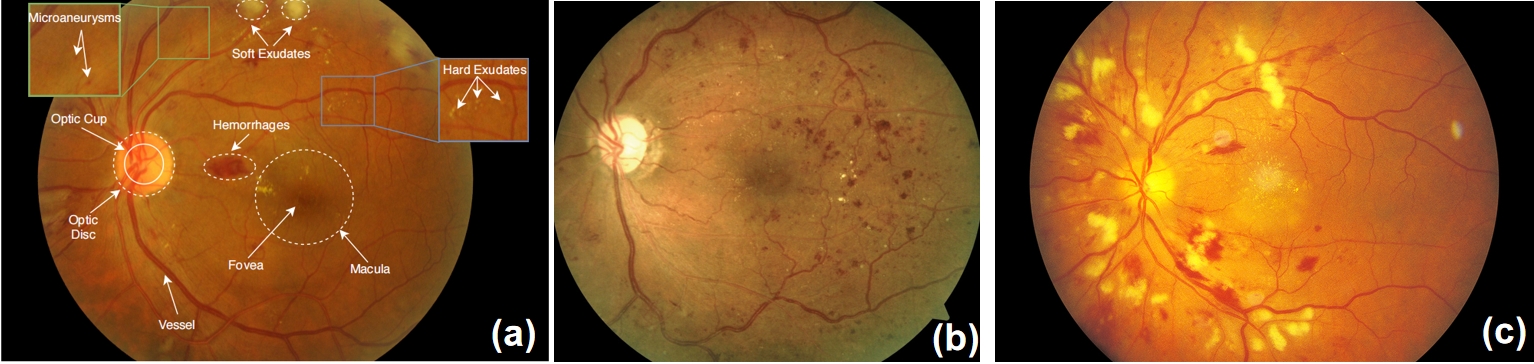}
\caption{(a) A fundus image from IDRiD dataset illustrating important biomarkers and lesions. (b) An illustration of Diabetic Retinopathy fundus image. (c) An illustration of Hypertensive Retinopathy fundus image.}
\label{background}
\end{figure}

Retinal vessel segmentation is crucial for preventing, diagnosing and assessing ocular diseases, which often manifest through changes in retinal vascular morphology. Conditions such as diabetes and hypertension significantly affect the appearance of retinal blood vessels. Diabetic Retinopathy (DR), a complication of diabetes, occurs when high blood sugar levels cause retinal vessels to leak and swell \cite{smart2015study} in Figure \ref{background}(b). Similarly, Hypertensive Retinopathy (HR), associated with high blood pressure, alters retinal vessels, making them twisted or narrower, indicating systemic pressure issues \cite{ding2014retinal} in in Figure \ref{background}(c). Therefore, analyzing retinal vessels from fundus images is a vital early detection method for several severe diseases.

Over the past few decades, research has focused on developing methods for retinal vessel segmentation, broadly categorized into manual and algorithm-based approaches. However, manual segmentation is challenging due to low contrast, complex structures and irregular illumination of retinal images, making it time-consuming and prone to errors \cite{na2018retinal,han2014blood}. Consequently, there is an urgent need for automated retinal vessel segmentation to alleviate the burden of manual analysis, improving both the speed and accuracy of retinal assessments.

With the advancements in machine learning and computer vision, deep learning frameworks have become competitive in capturing micro-vessels and providing detailed vascular features from retinal images, aiding clinicians in diagnosing and treating various eye diseases. Notably, Convolutional Neural Networks (CNNs) has offered robust feature representation in image classification and segmentation. Another significant advancement is the introduction of Fully Convolutional Networks (FCNs) \cite{long2015fully}, which pioneers end-to-end training for semantic segmentation. Drawing inspiration from FCNs, UNet \cite{ronneberger2015u} architecture ingeniously merges low-level features obtained from the analysis path with deeper features in the expansion path via encoder-decoder skip connections. This design enables the model to balance between capturing fine-grained local information and understanding broader contextual details, facilitating precise segmentation tasks. At present, numerous UNet variants have already been utilized for retinal vessel segmentation, including those with different convolution kernels (DUNet \cite{jin2019dunet}, DSCNet \cite{qi2023dynamic}), multiple interconnected networks (IterNet \cite{li2020iternet}, CTFNet \cite{wang2020ctf}), modified skip connections (BCDUNet \cite{azad2019bi}, AttUNet \cite{oktay2018attention}, UNet++ \cite{zhou2018unet++}) and models aiming to capture long-range dependencies (RollingUNet \cite{liu2024rolling}, MambaUNet \cite{wang2024mamba}). Furthermore, the development of Kolmogorov-Arnold Networks (KANs) \cite{liu2024kan} has offered superior interpretability and efficiency through the utilization of a series of nonlinear, learnable activation functions. Therefore, UKAN \cite{li2024u} integrates KANs into UNet framework, augmenting its capacity for non-linear modeling while also improving model interpretability.

However, retinal vessel segmentation still confronts several issues. The main objective of retinal vessel segmentation is to predict the class label for every pixel within a fundus image, with a primary focus on intra-image discrimination. Consequently, it is vital for models to extract discriminative and fine-grained pixel-level features as much as possible. Nevertheless, existing methods primarily focus on minimizing the difference between the output from decoder and label by utilizing various loss functions, but ignoring making full use of feature-level fine-grained representations from the encoder, which limits their performance on fine-grained retinal vessel segmentation. Fortunately, Contrastive Learning (CL) \cite{chen2020simple,he2020momentum} emerges as a promising approach to address these issues. As a subset of self-supervised learning, CL framework utilizes a suitable contrastive loss to efficiently pull similar representations closer together and push dissimilar representations apart, facilitating the extraction of more discriminative features. In general, considering the delicate and slender structure of retinal vasculature, it is extremely important to design a specific network and training strategy (loss function) to help the model extract more discriminative and fine-grained features, thereby achieving accurate vessel segmentation. And CL provides a promising solution to this problem.

In this work, to deal with the above issues, we propose a novel \textbf{Att}ention \textbf{U}-shaped \textbf{K}olmogorov-\textbf{A}rnold \textbf{N}etwork named \textbf{AttUKAN}, which incorporates Attention Gates (AGs) into UKAN to selectively filter features passed through skip connections. Additionally, we also design a novel loss named \textbf{L}abel-guided \textbf{P}ixel-wise \textbf{C}ontrastive \textbf{L}oss (\textbf{LPCL}) to supervise the model to extract more discriminative feature-level fine-grained representations. Beneficial from aforementioned two improvements, our model can maximize the use of feature-level fine-grained representations, hence guiding more precise retinal vessel segmentation. In summary, the main contributions of this paper are as follows:

\begin{itemize}
\item {A new retinal vessel segmentation model named AttUKAN is proposed to selectively filter skip connection features, thereby improving the performance of retinal vessel segmentation across various datasets.}
\item {A Label-guided Pixel-wise Contrastive Loss (LPCL) is designed to extract more discriminative features by distinguishing between foreground vessel-pixel sample pairs and background sample pairs.}
\item {State-of-the-art results have been achieved across four public datasets including DRIVE, STARE, CHASE-\text{\_}DB1, HRF and our private dataset for retinal vessel segmentation, compared with 11 networks. Comprehensive experiments and ablation studies are also conducted to verify the generalization ability and the effectiveness.}
\end{itemize}

\begin{table*}[t]
\centering
\setlength{\tabcolsep}{4.0pt}
\begin{tabular}{c|c|c}
\Xhline{1.0pt} 
Methods & Contribution & Limitation \\
\Xhline{1.0pt}
UNet \cite{ronneberger2015u} & encoder-decoder structure, skip connection & \multirow{9}{*}{only focus on minimizing} \\ 
DUNet \cite{jin2019dunet} & deformable convolution & \multirow{9}{*}{the difference between prediction} \\ 
DSCNet \cite{qi2023dynamic} & dynamic snake convolution & \multirow{9}{*}{from decoder and label} \\
IterNet  \cite{li2020iternet} & multiple iterations of miniUNet & \multirow{9}{*}{with various loss functions} \\
CTFNet \cite{wang2020ctf}  & coarse-to-fine networks & \\ 
BCDUNet \cite{azad2019bi} & BConvLSTM, densely convolutions & \\
AttUNet \cite{oktay2018attention} & attention gates & \\
UNet++ \cite{zhou2018unet++} & nested UNet structure \\
RollingUNet \cite{liu2024rolling} & CNN + MLP \\
MambaUNet \cite{wang2024mamba} & CNN + Mamba \\
UKAN \cite{li2024u} & CNN + KAN \\ \Xhline{1.0pt}
AttUKAN (Our) & extract discriminative fine-grained feature from encoder & - \\
\Xhline{1.0pt}
\end{tabular}
\caption{The key contribution of each method and our proposed AttUKAN for retinal vessel segmentation and the limitation of these baseline methods.}
\label{related work}
\end{table*}

\section{Related Work}
\label{sec2}

\subsection{Retinal Vessel Segmentation}
Early retinal vessel segmentation techniques were entirely unsupervised, utilizing standard image processing methods such as filtering, threshold segmentation, mathematical morphology and edge detection. For instance, \cite{singh2015local} proposed a method based on local entropy threshold segmentation. \cite{zana2001segmentation} proposed an algorithm grounded in mathematical morphology and curvature evaluation for the detection of vessel-like patterns in noisy environments. \cite{oliveira2016unsupervised} used a combined matched filter, Frangi’s filter and Gabor Wavelet filter to enhance the vessels. However, these methods still heavily relied on manually designed features and rules, limiting their flexibility and often failing to deliver optimal outcomes in complex scenarios.

The introduction of deep learning techniques represented a pivotal change in the field of retinal vessel segmentation, offering more sophisticated and accurate methods compared to traditional approaches. Recent research has explored vessel segmentation challenges through deep learning methodologies. For example, DeepVessel \cite{fu2016deepvessel} utilized a multi-scale, multi-level network coupled with a lateral output layer for retinal vessel segmentation, aiming to capture complex pixel interactions through conditional random fields. Bidirectional Symmetric Cascade Network (BSCN) \cite{guo2020bscn} innovated by incorporating dense dilated layers that dynamically adjusted their dilation rate according to vessel thickness, enhancing the segmentation of retinal vessels across varying scales. Vessel graph network (VGN) \cite{shin2019deep} enhanced their model by integrating convolutional and graph-convolutional layers, aiming to understand and represent the global connections within vessels more effectively.

Besides, the advancement of Convolutional Neural Networks (CNNs) enabled automated feature collection from images, eliminating the need for manual feature engineering. As shown in Table \ref{related work}, UNet \cite{ronneberger2015u} stood out for its effective use of encoder-decoder structure and skip-connection, enabling precise delineation of anatomical structures. 
Hence, numerous UNet variant networks were utilized for retinal vessel segmentation tasks. For instance, models were designed for different types of convolution kernels. Specifically, DUNet \cite{jin2019dunet} integrated deformable convolution and DSCNet \cite{qi2023dynamic} introduced dynamic snake convolution to enhance the segmentation of tubular structures. Furthermore, some models were designed to incorporate multiple networks. IterNet \cite{li2020iternet} utilized multiple iterations of miniUNet to enhance vessel details, while CTFNet \cite{wang2020ctf} employed a coarse-to-fine supervision strategy. Alternatively, several models modified skip connections. BCDUNet \cite{azad2019bi} utilized the strengths of both BConvLSTM states on skip connections and densely connected convolutions. AttUNet \cite{oktay2018attention} incorporated Attention Gates into skip connections and augmented predictive accuracy and sensitivity by attenuating and suppressing irrelevant feature activations.
UNet++ \cite{zhou2018unet++} introduced a nested UNet architecture that enhanced segmentation accuracy. Moreover, others were designed to capture long-range dependencies. RollingUNet \cite{liu2024rolling} introduced a module combining CNN and MLP and MambaUNet \cite{wang2024mamba} merged Mamba architecture to enhance spatial information transfer across scales. Furthermore, the development of Kolmogorov-Arnold Networks (KANs) \cite{liu2024kan} offered superior interpretability and efficiency through the utilization of a series of nonlinear, learnable activation functions. Therefore, UKAN \cite{li2024u} integrated KANs into UNet framework, augmenting its capacity for non-linear modeling while also improving model interpretability.
Despite the advancements in techniques like UNet, challenges still remain in segmenting retinal vessels accurately. A significant reason was that existing methods ignored the full utilization of feature-level fine-grained representations from the encoder, consequently failing to adequately focus on discriminative analysis within images.

\subsection{Contrastive Learning}

In recent years, Contrastive Learning (CL) has demonstrated remarkable success in acquiring discriminative features with a small number of annotations, significantly cutting down on the costs associated with manual annotation. CL aimed to draw similar representations closer together and separate dissimilar representations apart, by constructing positive and negative sample pairs. Recently, CL has been widely employed in self-supervised representation learning. For instance, MoCo \cite{he2020momentum} utilized a dynamic dictionary constructed with a queue and a moving-averaged encoder to store and compare image features. SimCLR \cite{chen2020simple} leveraged a large batch size to ensure a substantial presence of negative samples within each iteration, thereby facilitating their concurrent processing.

Rather than directly applying a CL framework, we construct a novel contrastive loss to guide the model in extracting more discriminative feature-level fine-grained representations. To the best of our knowledge, the work presented in this paper marks the first attempt to enhance retinal vessel segmentation by incorporating CL at the feature level to get more discriminative representations.

\begin{figure*}[t]
\centering
\includegraphics[width=\textwidth]{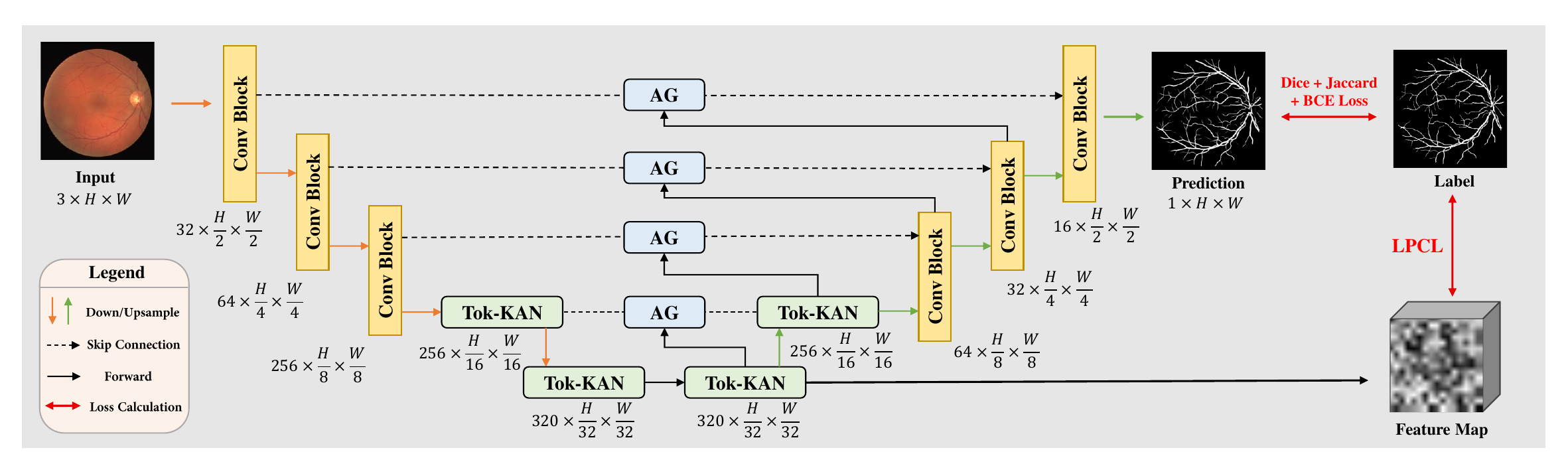}
\caption{Overview of AttUKAN pipeline. the input retinal vessel image is processed through the encoder, consisting of three convolutional blocks and two tokenized KAN blocks and the symmetric decoder. Attention Gates are incorporated into skip connection and the model is optimized with a hybrid baseline loss along with our proposed LPCL loss.}
\label{fig1}
\end{figure*}

\section{Method}\label{sec3}
This section focuses on introducing our proposed method, including a novel retinal vessel segmentation network: AttUKAN and a novel contrastive loss: LPCL. Firstly, a brief overview of retinal vessel segmentation is provided in Section \ref{sec3.1}. Secondly, a preliminary of KAN is introduced in Section \ref{sec.3.2}. Thirdly, the principles and architecture of AttUKAN is elaborated in Section \ref{sec3.3}. Finally, the designed LPCL is discussed in Section \ref{sec3.4}.

\subsection{Overview of Retinal Vessel Segmentation}\label{sec3.1}
The flowchart illustrating our proposed AttUKAN is shown in Figure \ref{fig1}. Given an input retinal fundus image $\boldsymbol{X}^0\in \mathbb{R}^{C^0 \times H \times W}$, where $H \times W$ signifies the spatial resolution of the image and $C^0$ denotes the number of channels, retinal vessel segmentation task aims to generate the corresponding pixel-wise semantic label map, matching the dimensions of $H \times W$ . To accomplish this objective, the segmentation network necessitates an encoder $e(\cdot)$ to extract discriminative features 
% $\left\{\boldsymbol{X}^1, \boldsymbol{X}^2, \boldsymbol{X}^3,\boldsymbol{X}^4,\boldsymbol{X}^5\right\}$ 
$\left\{\boldsymbol{X}^1, \cdots, \boldsymbol{X}^L\right\}$
from the input data. Then, a decoder $d(\cdot)$ is employed to integrate these features into $\boldsymbol{Y} \in \mathbb{R}^{ C_Y \times H \times W}$ to restore image specifics:
\begin{equation}
\boldsymbol{Y}=d(e(\boldsymbol{X^0}))=d\left(\left\{\boldsymbol{X}^1, \cdots, \boldsymbol{X}^L\right\}\right)
\end{equation}
where $\boldsymbol{X}^ \ell \in \mathbb{R}^{C_\ell \times \frac{H}{2^\ell} \times \frac{W}{2^\ell}}$ denotes the $\ell_{t h}$-level feature, $\ell \in \{1,\cdots,L\}$, L denotes the number of encoder layers, which is 5 in our AttUKAN. 

To optimize our proposed AttUKAN, we utilize a hybrid loss as the baseline loss including: binary cross-entropy loss $\mathcal{L}_{BCE}$, jaccard loss $\mathcal{L}_{jaccard}$ and dice loss $\mathcal{L}_{dice}$, then along with our proposed Label-guided Pixel-wise Contrastive Loss $\mathcal{L}_{LPCL}$. Specifically, the hybrid loss can be formulated as:
\begin{equation}
\mathcal{L}_{all}=\lambda_1\mathcal{L}_{BCE}+
\lambda_2\mathcal{L}_{jaccard}+
\lambda_3 \mathcal{L}_{dice}+
\lambda_4\mathcal{L}_{LPCL}
\end{equation}
where $\lambda_1$, $\lambda_2$, $\lambda_3$ and $\lambda_4$ are the weighting coefficients respectively. The detailed exposition of the aforementioned loss functions will be delineated in Section \ref{sec3.4}.

\begin{figure*}[t]
\centering
\includegraphics[width=0.9\textwidth]{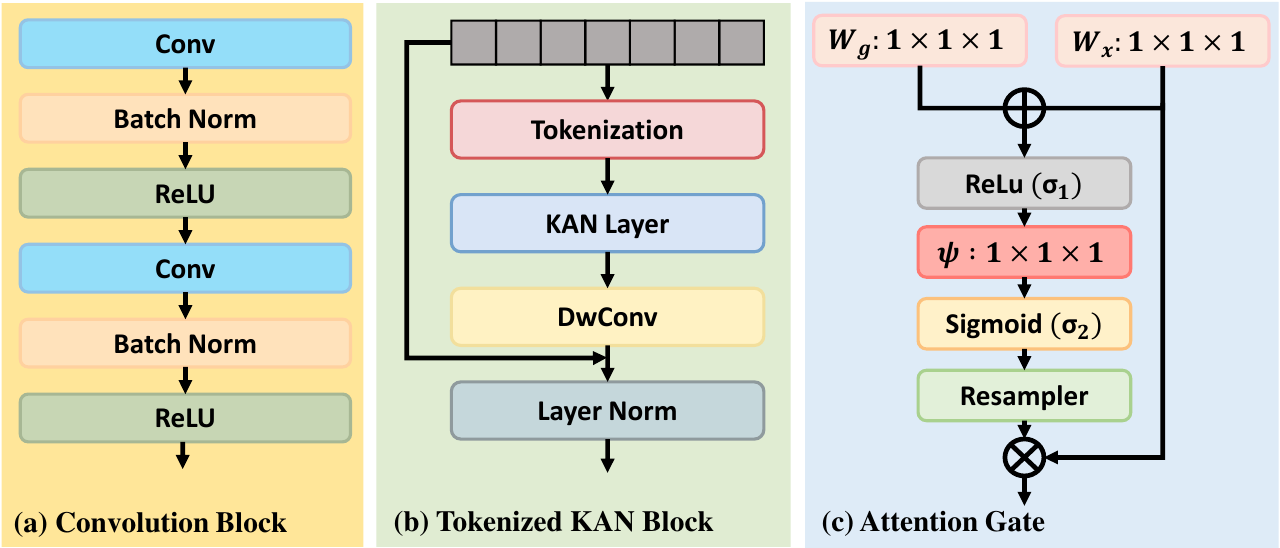}
\caption{Illustrations of each module in AttUKAN. (a) Convolution Block aims to extract features by leveraging convolutional operations. (b) Tokenized KAN Block implements KAN layers to augment the model's capacity for non-linear modeling. (c) Attention Gate aims to selectively filter and enhance feature-level fine-grained representations.}
\label{fig2}
\end{figure*}

\begin{figure}[t]
\centering
\includegraphics[width=0.45\textwidth]{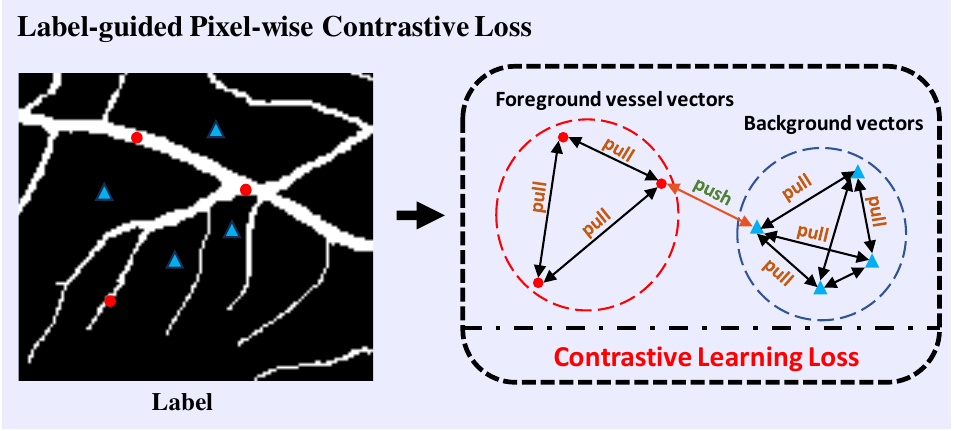}
\caption{Label-guided Pixel-wise Contrastive Loss. Red circles denote foreground vessel vectors, while blue triangles represent background vectors. Contrastive Learning Loss is employed to pull similar representations closer together and push dissimilar representations apart to supervise our AttUKAN learning more discriminative feature-level representation.}
\label{fig3}
\end{figure}

\subsection{Preliminary of KAN}\label{sec.3.2}
Recently, Multi-Layer Perceptron (MLP) is typically incorporated to model complex functional mappings by treating linear transformations and non-linearities separately. Specifically, MLP comprising L layers can be formulated as:
\begin{equation}
    \operatorname{MLP}(\boldsymbol{X}^0) = (W_{L-1} \operatorname{\circ} \sigma \operatorname{\circ} W_{L-2} \operatorname{\circ} \sigma \operatorname{\circ} \cdots \operatorname{\circ} W_1 \operatorname{\circ} \sigma \operatorname{\circ} W_0)\boldsymbol{X}^0
\end{equation}
where $W_i$ denotes transformation matrices, $i \in \{0,1,\cdots, L-1\}$,  $\sigma$ denotes activation functions and $\circ$ represents the composition of functions, meaning the output of one function becomes the input of the next.
However, the inherent complexity within this structure significantly hinders both model interpretability and parameter efficiency. Therefore, KANs \cite{liu2024kan} aim to address these issues by differentiating themselves through the use of learnable activation functions on the edges and parametrized activation functions as weights, thus eliminating the need for linear weight matrices. Specifically, KANs comprising L layers can be formulated as:
\begin{equation}
    \operatorname{KAN}(\boldsymbol{X}^0) = (\Phi_{L-1} \operatorname{\circ} \Phi_{L-2} \operatorname{\circ} ... \operatorname{\circ} \Phi_1 \operatorname{\circ} \Phi_0)\boldsymbol{X}^0
\end{equation}
where $\Phi$ comprises $n_{in} \times n_{out}$ learnable activation functions $\phi$, with $n_{in}$ - dimensional
input and $n_{out}$ - dimensional output for each KAN layer. Specifically, $\Phi$ can be formulated as:
\begin{equation}
    \Phi = \left\{\phi_{p, q}\right\} ,   p=1,2,\cdots,n_{in}, q=1,2,\cdots,n_{out}
\end{equation}
This design enables KANs to achieve superior performance with smaller model sizes, making them particularly suitable for small-scale medical image datasets, such as fundus vessel datasets. Furthermore, KAN exhibits superior accuracy compared to MLP, which can facilitate the precise segmentation of fine structures, such as retinal vasculature.

\subsection{AttUKAN architecture}\label{sec3.3}

In retinal vessel segmentation, previous U-shaped CNN models with MLP struggle with complex non-linear modeling, leading to suboptimal representation and segmentation of fine vascular structures. Kolmogorov-Arnold Networks (KANs) strive to emulate functional mappings through a sequence of nonlinear transformations across multiple layers. Through the use of learnable activation functions, they eliminate the need for linear weight matrices, thus enhancing the model's interpretability. However, a limitation of existing models is their insufficient use of feature-level fine-grained representations. Based on this premise, we propose a novel model named AttUKAN which adopts the conventional U-shaped encoder-decoder architecture while embedding KANs to augment its capacity for non-linear modeling. Additionally, we implement Attention Gates on skip connections to extract more feature-level fine-grained representations and enhance model sensitivity by suppressing irrelevant feature activations.

As shown in the Figure \ref{fig1}, the input image is processed through the encoder, consisting of three standard convolutional blocks and two tokenized KAN blocks, facilitating a deeper level of feature extraction. Alternatively, the decoder is symmetric to that of the encoder, enhancing feature refinement and reconstruction. Each encoder block reduces the feature resolution by half through down-sampling and each decoder block restores it by doubling through up-sampling.

\noindent{\textbf{Convolutional Block:}} As depicted in Figure \ref{fig2}(a), each convolution block within the model comprises three critical components: a convolutional layer (Conv), a batch normalization layer (Batch Norm) and a rectified linear unit (ReLU) activation function. Convolution blocks of the encoder incorporate 2x2 max-pooling, whereas convolution blocks of the decoder include bilinear interpolation for feature map upscaling. Formally, the output of each convolution block, can be articulated as:
\begin{equation}
\boldsymbol{X}^{\ell+1}=\operatorname{Pool}\left(\operatorname{Conv}\left(\boldsymbol{X}^{\ell}\right)\right)
\end{equation}
where $\boldsymbol{X}^\ell$ ($\ell$ = 0,1,2) represents the $\ell_{t h}$-level feature. 

\noindent{\textbf{Tokenized KAN Block:}} In the tokenized KAN block as shown in Figure \ref{fig2}(b), tokenization \cite{chen2024tokenunify,dosovitskiy2020image} is initiated by transforming the output feature from the convolutional layer into a series of 2D flattened patches. Each patch is then mapped into a latent D-dimensional embedding space through a trainable linear projection, effectively converting spatial features into a sequence of token. After tokenization, the feature tokens undergo processing through a sequence of three KAN layers $(N=3)$. 
Then, the features are refined via an efficient depth-wise convolution (DwConv) \cite{cao2022conv}, batch normalization and ReLU activation. Finally, a residual connection retains the original tokens, which are added back after these operations and then a layer normalization (LN) \cite{ba2016layer} follows. Specifically, the output of $\ell_{t h}$ tokenized KAN block can be formulated as:
\begin{equation}
\boldsymbol{X}^{\ell+1}=\operatorname{LN}\left(\boldsymbol{X}^{\ell}+\operatorname{DwConv}\left(\operatorname{KAN}\left(\boldsymbol{X}^{\ell}\right)\right)\right)
\end{equation}
where $\boldsymbol{X}^\ell$ ($\ell$ = 3,4) denotes the $\ell_{t h}$-level feature.  

\noindent{\textbf{Attention Gate:}} As shown in Figure \ref{fig2}(c), Attention Gates are incorporated within skip connections to facilitate communication between the encoder and decoder blocks. Therefore, to filter more feature-level fine-grained representations through skip connections. Initially, an attention value $\alpha^\ell\in \mathbb{R}^{H \times W}$ is computed for each input feature $\boldsymbol{X}^\ell$ and a gating feature $\boldsymbol{X}^{\ell+1}$ derived from a higher-level layer is used to determine focus regions. Then, additive attention \cite{bahdanau2014neural} $q_{a t t}^\ell\in \mathbb{R}^{H \times W}$ is used to obtain the gating coefficient, which can formulated as:
\begin{equation}
q_{a t t}^\ell=\psi^T\left(\operatorname{ReLU}\left(W_x^T \boldsymbol{X}^\ell+W_g^T \boldsymbol{X}^{\ell+1}+\boldsymbol{B}_g\right)\right)+b_\psi
\end{equation}
Then, a sigmoid function (Sigmoid) is incorporated, thus the attention value is formulated as follows:
\begin{equation}
\alpha^\ell=\operatorname{Sigmoid}\left(q_{att}^\ell\left(\boldsymbol{X}^\ell, \boldsymbol{X}^{\ell+1} ; \Theta_{att}\right)\right)
\end{equation}
where parameters $\Theta_{att}$ contain: linear transformations $W_x\in\mathbb{R}^{C_\ell \times C_{int}}$, $W_g\in\mathbb{R}^{C_{\ell+1} \times C_{int}}$, $\psi\in\mathbb{R}^{C_{int} \times 1}$ which are  computed using channel-wise 1x1x1 convolutions for the input tensors and bias terms $b_\psi\in \mathbb{R}^{H \times W}$, $\boldsymbol{B}_g\in \mathbb{R}^{C_{int} \times H \times W}$. $C_{int}$ represents a intermediate channel dimension.

Subsequently, attention coefficients are applied element-wise to the input feature maps $\hat{\boldsymbol{X}}^\ell=\boldsymbol{X}^\ell \cdot \alpha^\ell$. Finally, the output of the attention gates is concatenated with features from the last up-sampling block, which can be formulated as:
\begin{equation}
\boldsymbol{X}^{\ell}=\operatorname{Cat}\left(\boldsymbol{X}^{{\ell+1}},\left(\boldsymbol{\hat{X}}^{\ell}\right)\right)
\end{equation}
where $\boldsymbol{X}^\ell$ ($\ell$ = 1,2,3,4) represents the feature maps at $\ell_{t h}$-layer. 

\subsection{Label-guided Pixel-wise Contrastive Loss}\label{sec3.4}

Assured by the efficacy of our proposed AttUKAN architecture on extracting feature-level fine-grained representations delineated in Section \ref{sec3.3}, we further optimize our designed LPCL as shown in Figure \ref{fig3} in order to fully capitalize on the discriminative fine-grained features in our proposed model. Precisely, 
given a set of $N$ stochastic slice selections, $\left\{\boldsymbol{X_i}\right\}_{i=1 \ldots N}$ , the augmented mini-batch manifests as $\left\{\boldsymbol{\tilde{X}_i}\right\}_{i=1 \ldots 2 N}$ samples, wherein $\boldsymbol{\tilde{X}_{2 i}}$ and $\boldsymbol{\tilde{X}_{2 i-1}}$ are two random augmentations of $\boldsymbol{X_i}$. In this work, $\boldsymbol{X_i}$ is forwarded from the bottleneck of AttUKAN, specifically the $L_{th}$-level feature ${\boldsymbol{X}}^L$, thus forming $\boldsymbol{\tilde{X}_i^L}$, with a size of $S \times S$. $S$ denotes the spatial size of the feature map. For every two input images, we can form $S^2$ pixel-level contrastive pairs (positive pairs or negative pairs depending on whether these two pixels belongs to the same class).
Then, LPCL can be mathematically represented as:

\begin{equation}
\scalebox{0.92}{
$\displaystyle
\mathcal{L}_{\text{LPCL}}=\sum_{i=1}^{2 N}-\frac{1}{\left|\Omega_i^{+}\right|} \sum_{j \in \Omega_i^{+}} \frac{1}{S^2}\sum_{s} \log \frac{e^{\operatorname{sim}\left(\boldsymbol{\tilde{X}_{i,s}^L}, \boldsymbol{\tilde{X}_{j,s}^L}\right) / \tau}}{\sum_{k=1}^{2 N} \mathbb{I}_{i \neq k} \cdot e^{\operatorname{sim}\left(\boldsymbol{\tilde{X}_{i,s}^L}, \boldsymbol{\tilde{X}_{k,s}^L}\right) / \tau}}
$}
\end{equation}

where, ${\left|\Omega_i^{+}\right|}$ is set of indices denoting positive samples to $\boldsymbol{\tilde{X}_{i,s}}$. $\operatorname{sim}(\cdot, \cdot)$ embodying the cosine similarity metric, quantifies the resemblance between two vectors within the representation space. $\tau$ is a temperature scaling parameter and $\mathbb{I}$ is an indicator function.
With our proposed LPCL supervising AttUKAN, the model is capable of pulling intra-class pixels (\textbf{foreground:} vessel pixel $\leftrightarrow$ vessel pixel, \textbf{background:} non-vessel pixel $\leftrightarrow$ non-vessel pixel) together and pushing inter-class pixels (vessel pixel $\leftrightarrow$ non-vessel) apart in the feature space. Therefore, our AttUKAN can extract more discriminative fine-grained feature-level representations, hence achieving more accurate retinal vessel segmentation.

As detailed in Section \ref{sec3.1}, our model incorporates a hybrid loss function beyond the aforementioned strategies. The first term is a binary cross-entropy loss $\mathcal{L}_{BCE}$, designed to encourage the segmentation model to predict the correct class label at each pixel location independently and can be defined as: 
% \begin{equation}
% \mathcal{L}_{BCE}=-(\hat{\boldsymbol{Y}} \log \boldsymbol{Y}+\left(1-\hat{\boldsymbol{Y}}\right) \log \left(1-\boldsymbol{Y}\right))
% \end{equation}
\begin{equation}
\mathcal{L}_{\text{BCE}} = -\frac{1}{n} \sum_{i=1}^{n} \left( Y_i \log(\hat{Y}_i) + (1 - Y_i) \log(1 - \hat{Y}_i) \right)
\end{equation}

where $\hat{{Y}}$ represents the ground truth and ${Y}$ the predicted segmentation map. The second term,  jaccard loss $\mathcal{L}_{jaccard}$, can detect and correct higher-order inconsistencies between ground truth and predicted segmentation maps, can be formulated as:
% \begin{equation}
% \mathcal{L}_{jaccard}=1-\frac{|\hat{\boldsymbol{Y}} \cap \boldsymbol{Y}|}{|\hat{\boldsymbol{Y}} \cup \boldsymbol{Y}|}
% \end{equation}
\begin{equation}
\mathcal{L}_{\text{jaccard}} = 1 - \frac{\sum_{i=1}^{n} \hat{Y}_i Y_i}{\sum_{i=1}^{n} \hat{Y}_i + \sum_{i=1}^{n} Y_i - \sum_{i=1}^{n} \hat{Y}_i Y_i}
\end{equation}

The third term, dice loss $\mathcal{L}_{dice}$, is deployed to promote accurate pixel-wise class prediction and mitigate complex inconsistencies, can be articulated as:
% \begin{equation}
% \mathcal{L}_{dice}=1-\frac{2|\hat{\boldsymbol{Y}} \cap \boldsymbol{Y}|}{|\hat{\boldsymbol{Y}}|+|\boldsymbol{Y}|}
% \end{equation}
\begin{equation}
\mathcal{L}_{\text{dice}} = 1 - \frac{2 \sum_{i=1}^{n} \hat{Y}_i Y_i}{\sum_{i=1}^{n} \hat{Y}_i + \sum_{i=1}^{n} Y_i}
\end{equation}

Together with our proposed LPCL, these losses constitute our overall loss function $\mathcal{L}_{all}$, as shown in Section \ref{sec3.1}.

%DRIVE
\begin{table*}[ht]
\centering
\caption{Quantitative results on DRIVE dataset (best results shown in bold).}\label{tab1}
\footnotesize
\begin{tabular}{c|ccccccccccc}
\hline
\multirow{2}{*}{Methods} & \multicolumn{10}{c}{DRIVE} \\
& $\text {ACC (\%)}\uparrow$ & $\text {SE (\%)}\uparrow$ & $\text {SP (\%)}\uparrow$ &  $\text {F1 (\%)}\uparrow$  & $\text {MIoU (\%)}\uparrow$ & $\text {HD95}\downarrow$ &$\text {AUC (\%)}\uparrow$ & $\text {C (\%)}\uparrow$ & $\text {A (\%)}\uparrow$ &  $\text {L (\%)} \uparrow$ & $\text {F (\%)}\uparrow$ \\
\hline
UNet \cite{ronneberger2015u}
&$95.48$ &$77.51$ &$\mathbf{98.12}$ &$81.25$ &$68.48$ &$4.85$ &$97.64$ &$98.96$ &$93.81$ &$88.61$ &$82.34$ \\
DUNet \cite{jin2019dunet}
&$\mathbf{95.58}$ &$79.70$ &$97.92$ &$82.02$ &$69.56$ &$5.68$ &$97.93$ &$98.84$ &$93.48$ &$87.54$ &$80.99$ \\
DSCNet \cite{qi2023dynamic}
&$95.11$ &$83.09$ &$96.90$ &$81.14$ &$68.29$ &$6.34$ &$96.59$ &$98.66$ &$93.17$ &$86.72$ &$79.83$ \\

IterNet \cite{li2020iternet}
&$95.48$ &$78.34$ &$98.01$ &$81.44$ &$68.73$ &$4.22$ &$97.54$ &$\mathbf{99.01}$ &$94.33$ &$89.38$ &$83.56$ \\
CTFNet \cite{wang2020ctf} 
&$95.11$ &$82.28$ &$97.01$ &$80.98$ &$68.08$ &$4.91$ &$97.55$ &$98.74$ &$93.44$ &$88.42$ &$81.66$ \\

BCDUNet \cite{azad2019bi}
&$95.39$ &$77.85$ &$97.98$ &$81.04$ &$68.15$ &$4.69$ &$97.61$ &$98.97$ &$93.94$ &$88.77$ &$82.62$ \\
AttUNet \cite{oktay2018attention}
&$95.33$ &$83.42$ &$97.10$ &$81.88$ &$69.35$ &$4.30$ &$96.53$ &$98.77$ &$94.31$ &$89.4$ &$83.35$ \\
UNet++ \cite{zhou2018unet++}
&$95.50$ &$82.51$ &$97.43$ &$82.28$ &$69.92$ &$4.45$ &$97.79$ &$98.73$ &$94.14$ &$89.06$ &$82.85$ \\

RollingUNet \cite{liu2024rolling}
&$95.49$ &$80.03$ &$97.78$ &$81.78$ &$69.22$ &$4.51$ &$97.84$ &$98.80$ &$94.14$ &$89.11$ &$82.96$ \\
MambaUNet \cite{wang2024mamba}
&$95.51$ &$80.48$ &$97.73$ &$81.92$ &$69.41$ &$5.57$ &$97.77$&$98.73$ &$93.40$ &$87.37$ &$80.68$ \\
UKAN \cite{li2024u}
&$95.43$ &$82.96$ &$97.28$ &$82.13$ &$69.70$ &$5.40$ &$97.79$&$98.59$ &$93.83$ &$88.16$ &$81.65$ \\
AttUKAN (Ours)
&$95.49$ &$\mathbf{83.93}$ &$97.21$ &$\mathbf{82.50}$ &$\mathbf{70.24}$ &$\mathbf{4.21}$ &$\mathbf{97.95}$ &$98.66$ &$\mathbf{94.60}$ &$\mathbf{89.72}$ &$\mathbf{83.82}$ \\
\hline
\end{tabular}
\end{table*}

%STARE
\begin{table*}[ht]
\centering
\footnotesize
\caption{Quantitative results on STARE dataset (best results shown in bold).}\label{tab2}
\begin{tabular}{c|ccccccccccc}
\hline
\multirow{2}{*}{Methods} & \multicolumn{10}{c}{STARE} \\
& $\text {ACC (\%)}\uparrow$ & $\text {SE (\%)}\uparrow$ & $\text {SP (\%)}\uparrow$ &  $\text {F1 (\%)}\uparrow$  & $\text {MIoU (\%)}\uparrow$ & $\text {HD95}\downarrow$ &$\text {AUC (\%)}\uparrow$ & $\text {C (\%)}\uparrow$ & $\text {A (\%)}\uparrow$ &  $\text {L (\%)} \uparrow$ & $\text {F (\%)}\uparrow$ \\
\hline
UNet \cite{ronneberger2015u}
&$96.20$ &$70.71$ &$\mathbf{99.12}$ &$78.74$ &$65.49$ &$8.22$ &$97.62$ &$99.56$ &$88.59$ &$87.45$ &$77.51$ \\
DUNet \cite{jin2019dunet}  
&$96.17$ &$73.70$ &$98.81$ &$79.29$ &$66.23$ &$8.90$ &$97.80$ &$99.57$ &$88.91$ &$86.68$ &$77.08$ \\
DSCNet \cite{qi2023dynamic} 
&$95.85$ &$\mathbf{77.84}$ &$97.97$ &$79.15$ &$65.73$ &$9.97$ &$96.47$ &$99.78$ &$88.84$ &$86.69$ &$76.99$ \\
IterNet \cite{li2020iternet} 
&$96.20$ &$71.29$ &$99.06$ &$78.95$ &$65.70$ &$7.92$ &$97.47$ &$99.50$ &$89.33$ &$88.13$ &$78.64$ \\
CTFNet \cite{wang2020ctf} 
&$95.92$ &$73.20$ &$98.45$ &$77.93$ &$64.41$ &$9.76$ &$97.26$ &$99.72$ &$87.36$ &$86.19$ &$75.55$ \\
BCDUNet \cite{azad2019bi} 
&$96.20$ &$72.57$ &$98.91$ &$79.19$ &$66.05$ &$8.00$ &$97.63$ &$99.63$ &$89.16$ &$87.67$ &$78.19$ \\
AttUNet \cite{oktay2018attention} 
&$96.41$ &$76.20$ &$98.74$ &$80.99$ &$68.37$ &$\mathbf{7.79}$ &$96.67$ &$99.83$ &$90.38$ &$88.17$ &$79.76$ \\
UNet++ \cite{zhou2018unet++}
&$96.28$ &$76.83$ &$98.51$ &$80.36$ &$67.63$ &$9.04$ &$97.42$ &$99.84$ &$89.53$ &$87.83$ &$78.82$ \\
RollingUNet \cite{liu2024rolling}
&$96.21$ &$74.01$ &$98.76$ &$79.52$ &$66.48$ &$10.71$ &$97.62$ &$99.82$ &$88.74$ &$86.66$ &$77.12$ \\
MambaUNet \cite{wang2024mamba}
&$96.11$ &$73.19$ &$98.75$ &$78.96$ &$65.66$ &$9.23$ &$97.87$ &$99.77$ &$88.69$ &$86.09$ &$76.41$ \\
UKAN \cite{li2024u}
&$96.17$ &$73.58$ &$98.79$ &$79.26$ &$66.13$ &$10.81$ &$97.73$ &$\mathbf{99.87}$ &$87.87$ &$85.84$ &$75.69$ \\
AttUKAN (Ours)
&$\mathbf{96.43}$ &$76.88$ &$98.68$ &$\mathbf{81.14}$ &$\mathbf{68.64}$ &$7.98$ &$\mathbf{97.96}$ &$99.84$ &$\mathbf{90.65}$ &$\mathbf{88.33}$ &$\mathbf{79.97}$ \\
\hline
\end{tabular}
\end{table*}

%CHASE_DB1
\begin{table*}[ht]
\centering
\footnotesize
\caption{Quantitative results on CHASE$\_$DB1 dataset (best results shown in bold).}\label{tab3}
\begin{tabular}{c|ccccccccccc}
\hline
\multirow{2}{*}{Methods} & \multicolumn{10}{c}{CHASE$\_$DB1} \\
& $\text {ACC (\%)}\uparrow$ & $\text {SE (\%)}\uparrow$ & $\text {SP (\%)}\uparrow$ &  $\text {F1 (\%)}\uparrow$  & $\text {MIoU (\%)}\uparrow$ & $\text {HD95}\downarrow$ &$\text {AUC (\%)}\uparrow$ & $\text {C (\%)}\uparrow$ & $\text {A (\%)}\uparrow$ &  $\text {L (\%)} \uparrow$ & $\text {F (\%)}\uparrow$ \\
\hline
UNet \cite{ronneberger2015u}
&$96.27$ &$76.91$ &$98.47$ &$80.48$ &$67.38$ &$13.81$ &$98.04$ &$99.63$ &$89.57$ &$85.93$ &$76.71$ \\
DUNet \cite{jin2019dunet}  
&$95.70$ &$76.25$ &$97.90$ &$77.94$ &$63.90$ &$23.08$ &$96.32$ &$\mathbf{99.74}$ &$86.21$ &$82.37$ &$70.89$ \\
DSCNet \cite{qi2023dynamic} 
&$95.58$ &$79.75$ &$97.67$ &$79.28$ &$65.72$ &$20.73$ &$96.56$ &$99.56$ &$87.31$ &$83.07$ &$72.26$ \\
IterNet \cite{li2020iternet} 
&$\mathbf{96.34}$ &$77.11$ &$98.52$ &$80.80$ &$67.83$ &$14.09$ &$98.16$ &$99.65$ &$89.90$ &$86.03$ &$77.10$ \\
CTFNet \cite{wang2020ctf}  
&$95.49$ &$63.31$ &$\mathbf{99.10}$ &$73.65$ &$58.38$ &$49.30$ &$97.34$ &$99.58$ &$77.91$ &$77.91$ &$55.09$ \\
BCDUNet \cite{azad2019bi} 
&$96.14$ &$76.74$ &$98.34$ &$79.88$ &$66.54$ &$14.43$ &$97.92$ &$99.64$ &$89.07$ &$85.49$ &$75.90$ \\
AttUNet \cite{oktay2018attention} 
&$96.30$ &$77.04$ &$98.49$ &$80.60$ &$67.58$ &$13.43$ &$97.60$ &$99.67$ &$89.89$ &$86.31$ &$77.36$ \\

UNet++ \cite{zhou2018unet++}
&$96.16$ &$79.18$ &$98.09$ &$80.46$ &$67.36$ &$12.75$ &$97.96$ &$99.62$ &$89.81$ &$86.92$ &$77.80$ \\

RollingUNet \cite{liu2024rolling}
&$96.01$ &$80.00$ &$97.94$ &$80.35$ &$67.22$ &$14.79$ &$97.83$ &$99.60$ &$89.38$ &$86.22$ &$76.78$ \\
MambaUNet \cite{wang2024mamba}
&$95.44$ &$72.48$ &$98.03$ &$76.03$ &$61.38$ &$23.41$ &$96.25$ &$99.61$ &$84.03$ &$81.05$ &$67.89$ \\
UKAN \cite{li2024u}  
&$96.20$ &$75.92$ &$98.50$ &$79.96$ &$66.62$ &$16.10$ &$97.84$ &$99.56$ &$88.65$ &$85.26$ &$75.28$ \\
AttUKAN (Ours)
&$96.31$ &$\mathbf{80.61}$ &$98.11$ &$\mathbf{81.34}$ &$\mathbf{68.59}$ &$\mathbf{12.70}$ &$\mathbf{98.21}$ &$99.58$ &$\mathbf{90.39}$ &$\mathbf{87.06}$ &$\mathbf{78.38}$ \\
\hline
\end{tabular}
\end{table*}

%HRF
\begin{table*}[t]
\centering
\footnotesize
\caption{Quantitative results on HRF dataset (best results shown in bold).}\label{tab4}
\begin{tabular}{c|ccccccccccc}
\hline
\multirow{2}{*}{Methods} & \multicolumn{10}{c}{HRF} \\
& $\text {ACC (\%)}\uparrow$ & $\text {SE (\%)}\uparrow$ & $\text {SP (\%)}\uparrow$ &  $\text {F1 (\%)}\uparrow$  & $\text {MIoU (\%)}\uparrow$ & $\text {HD95}\downarrow$ &$\text {AUC (\%)}\uparrow$ & $\text {C (\%)}\uparrow$ & $\text {A (\%)}\uparrow$ &  $\text {L (\%)} \uparrow$ & $\text {F (\%)}\uparrow$ \\
\hline
UNet \cite{ronneberger2015u}
&$96.45$ &$74.59$ &$98.64$ &$79.19$ &$65.79$ &$35.73$ &$97.97$ &$99.96$ &$84.39$ &$82.82$ &$70.21$ \\
DUNet \cite{jin2019dunet}  
&$\mathbf{96.46}$ &$76.18$ &$98.49$ &$79.57$ &$66.35$ &$33.97$ &$\mathbf{97.99}$ &$99.91$ &$84.91$ &$82.82$ &$70.61$ \\
DSCNet \cite{qi2023dynamic} 
&$96.21$ &$76.73$ &$98.17$ &$78.67$ &$65.13$ &$38.27$ &$95.98$ &$99.95$ &$83.43$ &$81.65$ &$68.35$ \\
IterNet \cite{li2020iternet} 
&$96.39$ &$73.86$ &$\mathbf{98.66}$ &$78.82$ &$65.25$ &$35.87$ &$97.74$ &$99.95$ &$83.97$ &$81.85$ &$68.97$ \\
CTFNet \cite{wang2020ctf}  
&$95.49$ &$67.16$ &$98.35$ &$73.01$ &$57.86$ &$54.31$ &$95.97$ &$99.86$ &$74.75$ &$72.46$ &$54.52$ \\
BCDUNet \cite{azad2019bi} 
&$96.36$ &$75.10$ &$98.50$ &$78.93$ &$65.43$ &$37.56$ &$97.83$ &$99.96$ &$84.14$ &$81.80$ &$69.12$ \\
AttUNet \cite{oktay2018attention} 
&$96.42$ &$75.51$ &$98.03$ &$79.34$ &$65.96$ &$33.23$ &$97.80$ &$99.94$ &$84.93$ &$83.22$ &$70.90$ \\

UNet++ \cite{zhou2018unet++}
&$96.39$ &$\mathbf{79.81}$ &$98.06$ &$80.13$ &$67.09$ &$29.82$ &$97.52$ &$99.97$ &$86.02$ &$85.12$ &$73.39$ \\

RollingUNet \cite{liu2024rolling}
&$96.33$ &$79.50$ &$98.03$ &$79.84$ &$66.67$ &$31.69$ &$97.28$ &$99.96$ &$85.55$ &$84.56$ &$72.50$ \\
MambaUNet \cite{wang2024mamba}
&$96.12$ &$74.95$ &$98.27$ &$77.90$ &$64.08$ &$41.59$ &$96.38$ &$99.93$ &$82.43$ &$80.95$ &$66.95$ \\

UKAN \cite{li2024u}  
&$96.40$ &$79.66$ &$98.09$ &$80.14$ &$67.11$ &$27.87$ &$97.59$ &$\mathbf{99.97}$ &$86.11$ &$\mathbf{85.46}$ &$73.76$ \\

AttUKAN (Ours)
&$96.45$ &$79.02$ &$98.17$ &$\mathbf{80.21}$ &$\mathbf{67.21}$ &$\mathbf{26.92}$ &$97.89$ &$99.97$ &$\mathbf{86.30}$ &$85.40$ &$\mathbf{73.91}$ \\
\hline
\end{tabular}
\end{table*}

\section{EXPERIMENTAL RESULTS}\label{sec4}
\subsection{Datasets}\label{sec4.1}
Performance is evaluated on four public datasets, including DRIVE \cite{staal2004ridge}, STARE \cite{hoover2000locating}, CHASE\text{\_}DB1 \cite{owen2009measuring} and HRF \cite{odstrcilik2013retinal}, as well as our private dataset in the experiments. The DRIVE dataset comprises 40 color fundus photographs, including 20 training images and 20 testing images, with a resolution of 565 $\times$ 584 pixels. The STARE dataset comprises 20 fundus images with a resolution of 700 $\times$ 605 pixels, intended to assist ophthalmologists in diagnosing eye diseases. The CHASE\text{\_}DB1 dataset comprises 28 retinal images, taken from both eyes of 14 children, with a resolution of 999 $\times$ 960 pixels. The HRF dataset consists of 45 images, of which 15 are healthy, 15 have diabetic retinopathy and 15 are glaucomatous. Our private dataset \footnote{Our private dataset will be public in the future.} contains totally 115 gray images of both eyes captured at 548nm wavelengths from 60 patients. Notably, there are 5 patients with only one fundus image, either OD (Oculus Dexter/Right Eye) or OS (Oculus Sinister/Left Eye).
All the manual annotations of our private dataset were qualified by ophthalmology experts. All procedures involved in this private dataset were reviewed and approved by the Ethics Committee of Peking University Health Science Center (PUIRB-YS2023166). Consent was given for publication by the participants. And the research was conducted in accordance with the principles embodied in the Declaration of Helsinki and in accordance with local statutory requirements.
Compared to the other three datasets, images from HRF and our private dataset have higher resolutions of 3504 $\times$ 2336 pixels and 2730 $\times$ 2048 pixels, respectively.

Since no predefined splits for training or testing are provided for STARE/CHASE\_DB1, following the experimental setting of DUNet \cite{jin2019dunet}, we use the first 10/14 images for training and the remaining 10/14 for testing. For HRF, we use 30 images for training and 15 for testing. For our private dataset, we use 90 images for training and 25 for testing. In this work, data augmentation is employed in our experiments. Following the DUNet \cite{jin2019dunet} setup, the original RGB images are converted into single-channel representations, then normalization and Contrast Limited Adaptive Histogram Equalization (CLAHE) and gamma correction are applied to the entire dataset.

\subsection{Metrics}\label{sec4.2}
The performance of our model is assessed using a comprehensive set of evaluation metrics, encompassing accuracy (ACC), sensitivity (SE), specificity (SP),  F1 score (F1), Mean Intersection over Union (MIoU), Hausdorff distance(95\%) (HD95) and the area under the receiver operating characteristic curve (AUC). Additionally, connectivity (C), overlapping area (A), consistency of vessel length (L) and the overall metric (F) are also utilized to evaluate our model's performance.

The AUC curve typically refers to the Receiver Operating Characteristic (ROC) curve, which plots the true positive rate (sensitivity) against the false positive rate (1-specificity) at various threshold settings. HD95 measures the similarity between the ground truth and segmentation results by calculating the distances from each point in one set to the nearest point in the other set, which can be formulated as:

\begin{equation}
\scriptsize
\text{HD}_{95}(A, B) = \max\left\{\underset{a \in A}{\text{quantile}
}_{95}\underset{b \in B}{\min} \, d(a, b), \underset{b \in B}{\text{quantile}}_{95}\underset{a \in A}{\min} \, d(b, a))\right\}
\end{equation}
where $A$ and $B$ are point sets of ground truth and segmentation result, ${\text{quantile}}_{95}$ represents the 95th percentile of all distance values.  and $d(a,b)$ is the Euclidean distance between points $a$ and $b$.

For the other metrics, ACC, SE, SP,  F1, MIoU are defined as follows:

\begin{equation}
\mathrm{ACC}=\frac{\mathrm{TP}+\mathrm{TN}}{\mathrm{TP}+\mathrm{FP}+\mathrm{TN}+\mathrm{FN}}
\end{equation}

\begin{equation}
\mathrm{SE}=\frac{\mathrm{TP}}{\mathrm{TP}+\mathrm{FN}}
\end{equation}

\begin{equation}
\mathrm{SP}=\frac{\mathrm{TN}}{\mathrm{TN}+\mathrm{FP}}
\end{equation}

\begin{equation}
\mathrm{F1}=\frac{2\mathrm{TP}}{2\mathrm{TP}+\mathrm{FP}+\mathrm{FN}}
\end{equation}

\begin{equation}
\mathrm{MIoU}=\frac{2\mathrm{TP}}{\mathrm{TP}+\mathrm{FP}+\mathrm{FN}}
\end{equation}
where TP denotes true positive samples, TN denotes true negative samples, FP denotes false positive samples; FN denotes false negative samples.

In addition, \cite{gegundez2011function} has proposed metrics specifically designed for vessel segmentation, which are widely utilized to evaluate the connectivity (C), overlapping area (A) and consistency of vessel length (L) in the predicted vessels. Connectivity evaluates the degree of fragmentation in vascular segmentation by comparing the number of connected components to the total number of vessel pixels. Overlapping area, based on the Jaccard coefficient, evaluates the degree of overlap between the ground truth and the segmentation result. Consistency of vessel length measures the degree of coincidence between the ground truth and the segmentation result in terms of total length.Specifically, these metrics can be formulated as follows:
\begin{equation}
C(S, S_{G}) = 1 - \min\left(1, \frac{\left|{\#}_{C}(S_{G}) - \#_{C}(S)\right|}{\#(S_{G})}\right)
\end{equation}

\begin{equation}
A(S, S_{G}) = \frac{\#(\delta_{\alpha}(S) \cap S_{G}) \cup (S \cap \delta_{\alpha}(S_{G}))}{\#(S \cup S_{G})}
\end{equation}

\begin{equation}
L(S, S_{G}) = \frac{\#((\varphi(S) \cap \delta_{\beta}(S_{G})) \cup (\delta_{\beta}(S) \cap \varphi(S_{G})))}{\#(\varphi(S) \cup \varphi(S_{G}))}.
\end{equation}
where $S$ and $S_{G}$ denotes the segmentation result and ground truth respectively, $\#_{C}(\cdot)$ and $\#(\cdot)$ represents the number of connected components and the cardinality. $\delta_{\alpha}$ and $\delta_{\beta}$ denote morphological dilation using a disc of radius $\alpha$ and $\beta$ pixels, respectively. $\varphi$ denotes an homotopic skeletonization. In this work, $\alpha$ and $\beta$ are set to 2.
Finally, the overall metric (F) is defined as follows:

\begin{equation}
    F(C, A, L) = C \times A \times L
\end{equation}

%Our
\begin{table*}[ht]
\centering
\footnotesize
\caption{Quantitative results on our private dataset (best results shown in bold).}\label{tab5}
\begin{tabular}{c|ccccccccccc}
\hline
\multirow{2}{*}{Methods} & \multicolumn{10}{c}{Our Private Dataset} \\
& $\text {ACC (\%)}\uparrow$ & $\text {SE (\%)}\uparrow$ & $\text {SP (\%)}\uparrow$ &  $\text {F1 (\%)}\uparrow$  & $\text {MIoU (\%)}\uparrow$ & $\text {HD95}\downarrow$ &$\text {AUC (\%)}\uparrow$ & $\text {C (\%)}\uparrow$ & $\text {A (\%)}\uparrow$ &  $\text {L (\%)} \uparrow$ & $\text {F (\%)}\uparrow$ \\
\hline
UNet \cite{ronneberger2015u}
&$95.47$ &$78.54$ &$97.56$ &$79.59$ &$66.27$ &$19.21$ &$97.17$ &$99.97$ &$84.81$ &$88.95$ &$75.49$ \\
DUNet \cite{jin2019dunet}  
&$95.27$ &$71.08$ &$98.44$ &$77.16$ &$63.07$ &$30.00$ &$97.28$ &$99.95$ &$80.82$ &$85.08$ &$68.88$ \\

DSCNet \cite{qi2023dynamic} 
&$95.23$ &$78.46$ &$97.44$ &$78.71$ &$65.07$ &$19.56$ &$95.53$ &$99.96$ &$83.89$ &$88.33$ &$74.16$ \\
IterNet \cite{li2020iternet} 
&$95.56$ &$76.84$ &$98.04$ &$79.58$ &$66.29$ &$18.08$ &$97.67$ &$99.97$ &$84.71$ &$88.82$ &$75.31$ \\
CTFNet \cite{wang2020ctf}  
&$95.23$ &$73.55$ &$98.08$ &$77.61$ &$63.62$ &$28.37$ &$96.03$ &$99.96$ &$81.58$ &$85.80$ &$70.10$ \\

BCDUNet \cite{azad2019bi}
&$\mathbf{95.57}$ &$76.49$ &$\mathbf{98.96}$ &$79.54$ &$66.23$ &$17.76$ &$97.73$ &$99.97$ &$84.70$ &$88.80$ &$75.29$ \\

AttUNet \cite{oktay2018attention} 
&$95.27$ &$74.29$ &$98.01$ &$77.91$ &$64.02$ &$25.89$ &$96.40$ &$99.95$ &$82.24$ &$86.43$ &$71.17$ \\
UNet++ \cite{zhou2018unet++}
&$95.47$ &$78.54$ &$97.71$ &$79.59$ &$66.27$ &$19.21$ &$97.17$ &$99.97$ &$84.81$ &$88.95$ &$75.91$ \\

RollingUNet \cite{liu2024rolling}
&$95.41$ &$78.32$ &$97.67$ &$79.32$ &$65.91$ &$20.28$ &$96.57$ &$99.97$ &$84.30$ &$88.50$ &$74.67$ \\
MambaUNet \cite{wang2024mamba}
&$95.18$ &$74.97$ &$97.84$ &$77.77$ &$63.81$ &$26.99$ &$96.20$ &$99.95$ &$82.12$ &$86.05$ &$70.72$ \\

UKAN \cite{li2024u}  
&$95.49$ &$79.93$ &$97.56$ &$79.98$ &$66.8$ &$16.01$ &$97.47$ &$\mathbf{99.97}$ &$85.61$ &$89.75$ &$76.88$ \\

AttUKAN (Ours)
&$95.47$ &$\mathbf{80.80}$ &$97.41$ &$\mathbf{80.09}$ &$\mathbf{66.94}$ &$\mathbf{14.65}$ &$\mathbf{97.73}$ &$99.97$ &$\mathbf{85.97}$ &$\mathbf{90.18}$ &$\mathbf{77.55}$ \\
\hline
\end{tabular}
\end{table*}

%average
\begin{table*}[t]
\centering
\footnotesize
\caption{Averaged quantitative results across DRIVE, STARE, CHASE\_DB1, HRF and our private dataset (best results shown in bold).}\label{tab6}
\begin{tabular}{c|ccccccccccc}
\hline
\multirow{2}{*}{Methods} & \multicolumn{10}{c}{Averaged Results} \\
& $\text {ACC (\%)}\uparrow$ & $\text {SE (\%)}\uparrow$ & $\text {SP (\%)}\uparrow$ &  $\text {F1 (\%)}\uparrow$  & $\text {MIoU (\%)}\uparrow$ & $\text {HD95}\downarrow$ &$\text {AUC (\%)}\uparrow$ & $\text {C (\%)}\uparrow$ & $\text {A (\%)}\uparrow$ &  $\text {L (\%)} \uparrow$ & $\text {F (\%)}\uparrow$ \\
\hline
UNet \cite{ronneberger2015u}
&$95.97$ &$75.65$ &$98.38$ &$79.85$ &$66.68$ &$16.36$ &$97.69$ &$99.62$&$88.23$ &$86.75$ &$76.45$ \\
DUNet \cite{jin2019dunet}  
&$95.84$ &$75.38$ &$98.31$ &$79.20$ &$65.82$ &$20.33$ &$97.46$ &$99.6$&$86.87$ &$84.90$ &$73.69$ \\
DSCNet \cite{qi2023dynamic} 
&$95.60$ &$79.17$ &$97.63$ &$79.39$ &$65.99$ &$18.97$ &$96.23$ &$99.58$ &$87.33$ &$85.29$ &$74.32$ \\
IterNet \cite{li2020iternet} 
&$95.99$ &$75.49$ &$98.46$ &$79.92$ &$66.76$ &$16.04$ &$97.72$ &$99.62$&$88.45$ &$86.84$ &$76.72$ \\
CTFNet \cite{wang2020ctf}  
&$95.45$ &$71.90$ &$98.20$ &$76.64$ &$62.47$ &$29.33$ &$96.83$ &$99.57$&$83.01$ &$82.16$ &$67.38$ \\
BCDUNet \cite{azad2019bi}
&$95.93$ &$75.75$ &$\mathbf{98.54}$ &$79.72$ &$66.48$ &$16.49$ &$97.74$ &$\mathbf{99.63}$&$88.20$ &$86.51$ &$76.22$ \\

AttUNet \cite{oktay2018attention} 
&$95.95$ &$77.29$ &$98.07$ &$80.14$ &$67.06$ &$16.93$ &$97.00$ &$99.63$&$88.35$ &$86.71$ &$76.51$ \\

UNet++ \cite{zhou2018unet++}
&$95.96$ &$79.37$ &$97.96$ &$80.56$ &$67.65$ &$15.05$ &$97.57$ &$99.63$&$88.86$ &$87.58$ &$77.75$ \\

RollingUNet \cite{liu2024rolling}
&$95.89$ &$78.37$ &$98.04$ &$80.16$ &$67.10$ &$16.40$ &$97.43$ &$99.63$&$88.42$ &$87.01$ &$76.81$ \\
MambaUNet \cite{wang2024mamba}
&$95.67$ &$75.21$ &$98.12$ &$78.52$ &$64.87$ &$21.36$ &$96.89$ &$99.60$&$86.13$ &$84.30$ &$72.53$ \\

UKAN \cite{li2024u}  
&$95.94$ &$78.41$ &$98.04$ &$80.29$ &$67.27$ &$15.24$ &$97.68$ &$99.59$&$88.41$ &$86.89$ &$76.65$ \\

AttUKAN (Ours)
&$\mathbf{96.03}$ &$\mathbf{80.25}$ &$97.92$ &$\mathbf{81.06}$ &$\mathbf{68.32}$ &$\mathbf{13.29}$ &$\mathbf{97.95}$ &$99.60$&$\mathbf{89.58}$ &$\mathbf{88.14}$ &$\mathbf{78.73}$\\
\hline
\end{tabular}
\end{table*}

\subsection{Implementation details}\label{sec4.3}
Our model is trained on an NVIDIA GeForce RTX 2080 Ti GPU and is built using the PyTorch framework. We randomly extract patches of size $64\times64$ from the original images due to large size of fundus images. A ratio of 9:1 is set for the training set and the validation set, respectively. We set batch size to 25, learning rate to 0.003, total number of training epochs to 100 and use Adam as the optimizer. 
The weighting coefficients of our loss function $\lambda_1$, $\lambda_2$, $\lambda_3$ as mentioned in Section \ref{sec3.1}, are set to 0.8, 0.2 and 1.0, respectively according to the previous experience \cite{wang2019dual}, while the weighing coefficients of LPCL, $\lambda_4$ is set to 0.3 in DRIVE, STARE and CHASE datasets and to 0.5 in HRF and our private datasets. Finally, the probability maps predicted by the model are binarized using a threshold of 0.5 to obtain the final segmentation.

\subsection{Experimental results and analysis}
\noindent{\textbf{Quantitative Results:}}
To verify the performance of our proposed AttUKAN in retinal vessel segmentation, experiments are first conducted on the five datasets. We compare the results with other state-of-the-art methods, including UNet \cite{ronneberger2015u}, UNet++ \cite{zhou2018unet++}, UKAN \cite{li2024u}, DSCNet \cite{qi2023dynamic}, RollingUNet \cite{liu2024rolling}, MambaUNet \cite{wang2024mamba}, IterNet \cite{li2020iternet}, DUNet \cite{jin2019dunet}, CTFNet \cite{wang2020ctf}, BCDUNet \cite{azad2019bi} and AttUNet \cite{oktay2018attention}. All experiments across different methods utilize the same dataset settings and partitions. The quantitative metrics are listed in Table \ref{tab1}-\ref{tab5}. Overall, our proposed AttUKAN attains the highest scores on most of the considered metrics, notably performing exceptionally well in the F1 score and MIoU, which reflect the similarity and diversity of the testing datasets. Specifically, AttUKAN achieves F1 scores of 82.50\%, 81.14\%, 81.34\%, 80.21\% and 80.09\% on the DRIVE, STARE, CHASE$\_$DB1, HRF and our private dataset, respectively, along with MIoU scores of 70.24\%, 68.64\%, 68.59\%, 67.21\% and 66.94\%, which are the highest among all state-of-the-art methods. On the other hand, our proposed AttUKAN outperforms in F metric on all the datasets, this indicates that the connectivity and accuracy of vessel segmentation in AttUKAN have achieved better outcomes. However, AttUKAN only achieves the highest score in ACC metric on the STARE dataset. Since ACC measures the proportion of correctly classified pixels and retinal blood vessels typically occupy only a small portion of the entire fundus image. Therefore, ACC is relatively less important in evaluating vessel segmentation results compared to F1 and MIoU scores. Besides, our AttUKAN also shows outstanding performance in HD95 and AUC on most of the datasets. AttUKAN achieves HD95 scores of 12.7, 26.92 and 14.65 on the CHASE\_DB1, HRF and our private dataset, along with AUC scores of 97.95\%, 97.96\%, 98.21\% and 97.73\% on the DRIVE, STARE, CHASE\_DB1 and our private dataset. Moreover, we also show the averaged quantitative results across DRIVE, STARE, CHASE\_DB1, HRF and our private dataset in Table \ref{tab6} and our proposed AttUKAN achieves highest performance in almost all metrics including ACC, SE, F1, MIoU, HD95, AUC, A, L and F.

\begin{figure*}[ht]
\centering
\includegraphics[width=0.8\textwidth]{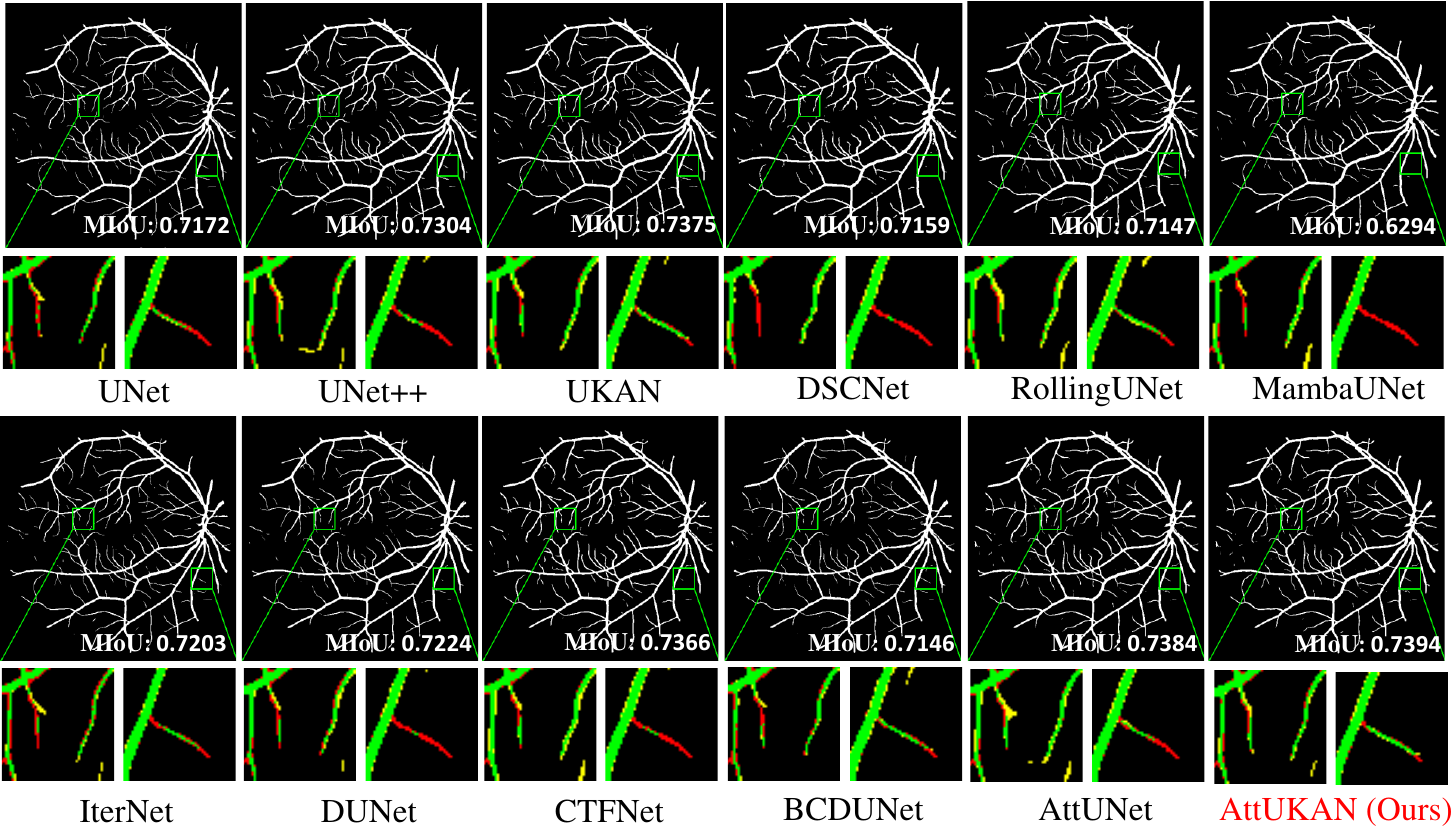}
\caption{Visualization of tiny vessel segmentation results on the DRIVE dataset. Green pixels represent true positive predictions; yellow pixels indicate false positive predictions; and red pixels denote false negatives predictions. Our proposed AttUKAN can get a more accurate segmentation result with more green regions and higher MIoU.}
\label{fig4}
\end{figure*}

\begin{figure*}[ht]
\centering
\includegraphics[width=0.8\textwidth]{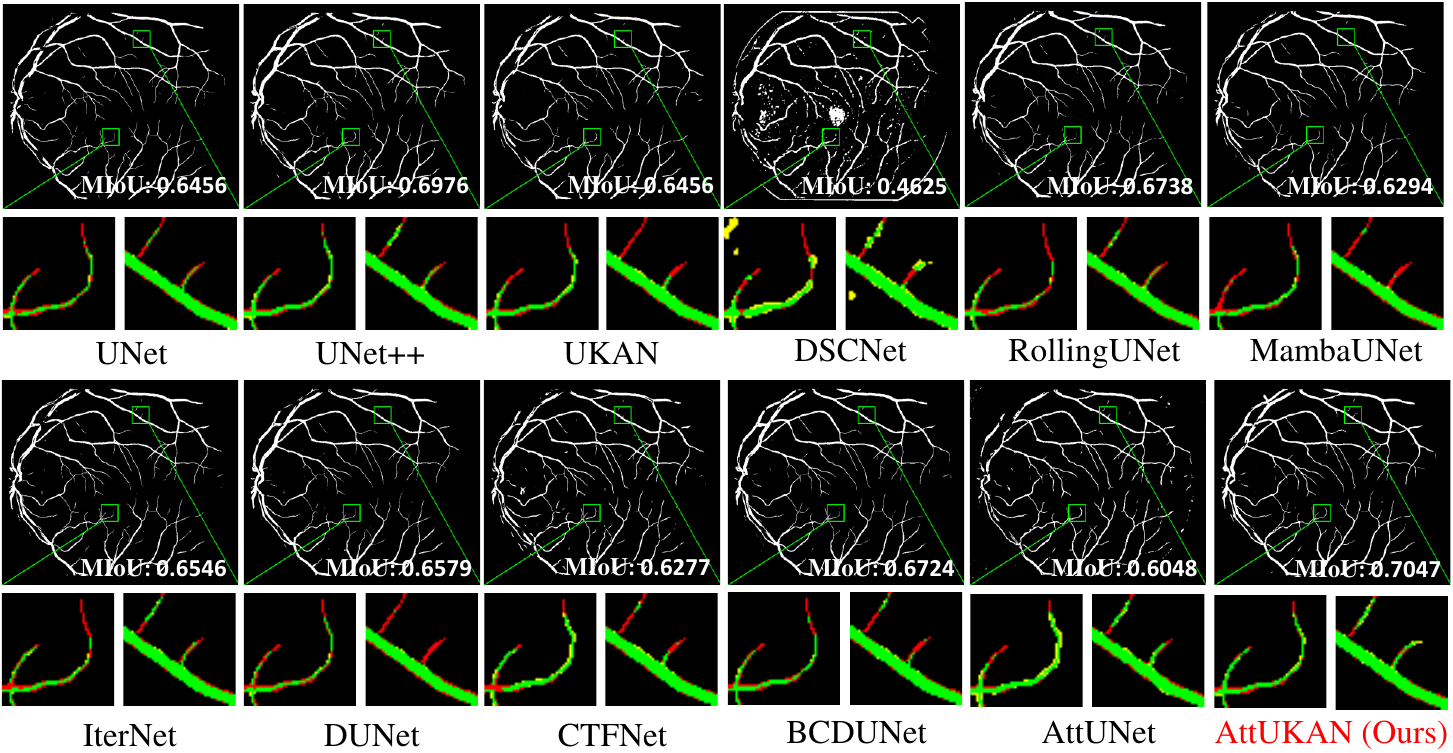}
\caption{Visualization of tiny vessel segmentation results on the STARE dataset. Green pixels represent true positive predictions; yellow pixels indicate false positive predictions; and red pixels denote false negatives predictions. Our proposed AttUKAN can get a more accurate segmentation result with more green regions and higher MIoU.}
\label{fig5}
\end{figure*}

\begin{figure*}[ht]
\centering
\includegraphics[width=0.8\textwidth]{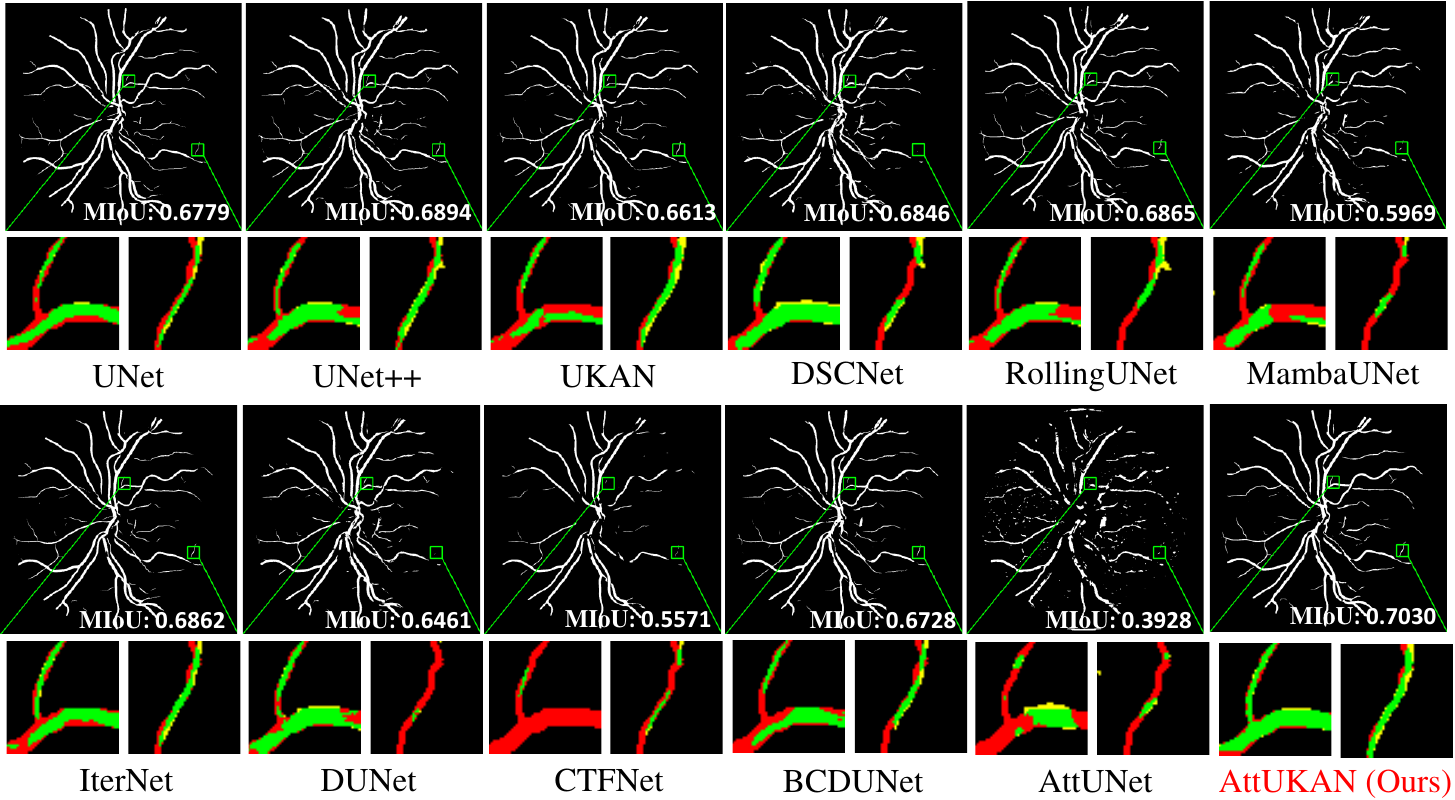}
\caption{Visualization of tiny vessel segmentation results on the CHASE$\_$DB1 dataset. Green pixels represent true positive predictions; yellow pixels indicate false positive predictions; and red pixels denote false negatives predictions. Our proposed AttUKAN can get a more accurate segmentation result with more green regions and higher MIoU.}
\label{fig6}
\end{figure*}

\begin{figure*}[ht]
\centering
\includegraphics[width=0.8\textwidth]{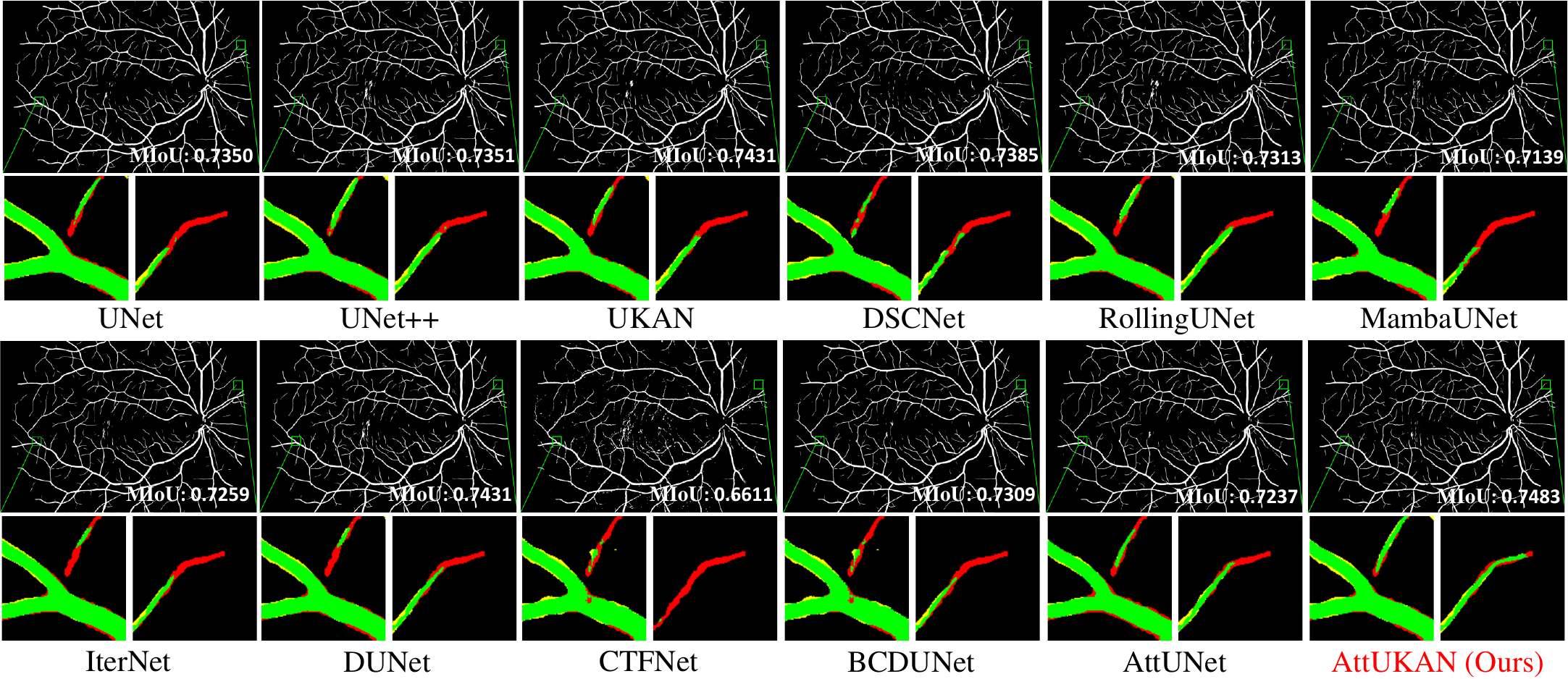}
\caption{Visualization of tiny vessel segmentation results on the HRF dataset. Green pixels represent true positive predictions; yellow pixels indicate false positive predictions; and red pixels denote false negatives predictions. Our proposed AttUKAN can get a more accurate segmentation result with more green regions and higher MIoU.}
\label{fig7}
\end{figure*}

\begin{figure*}[h]
\centering
\includegraphics[width=0.8\textwidth]{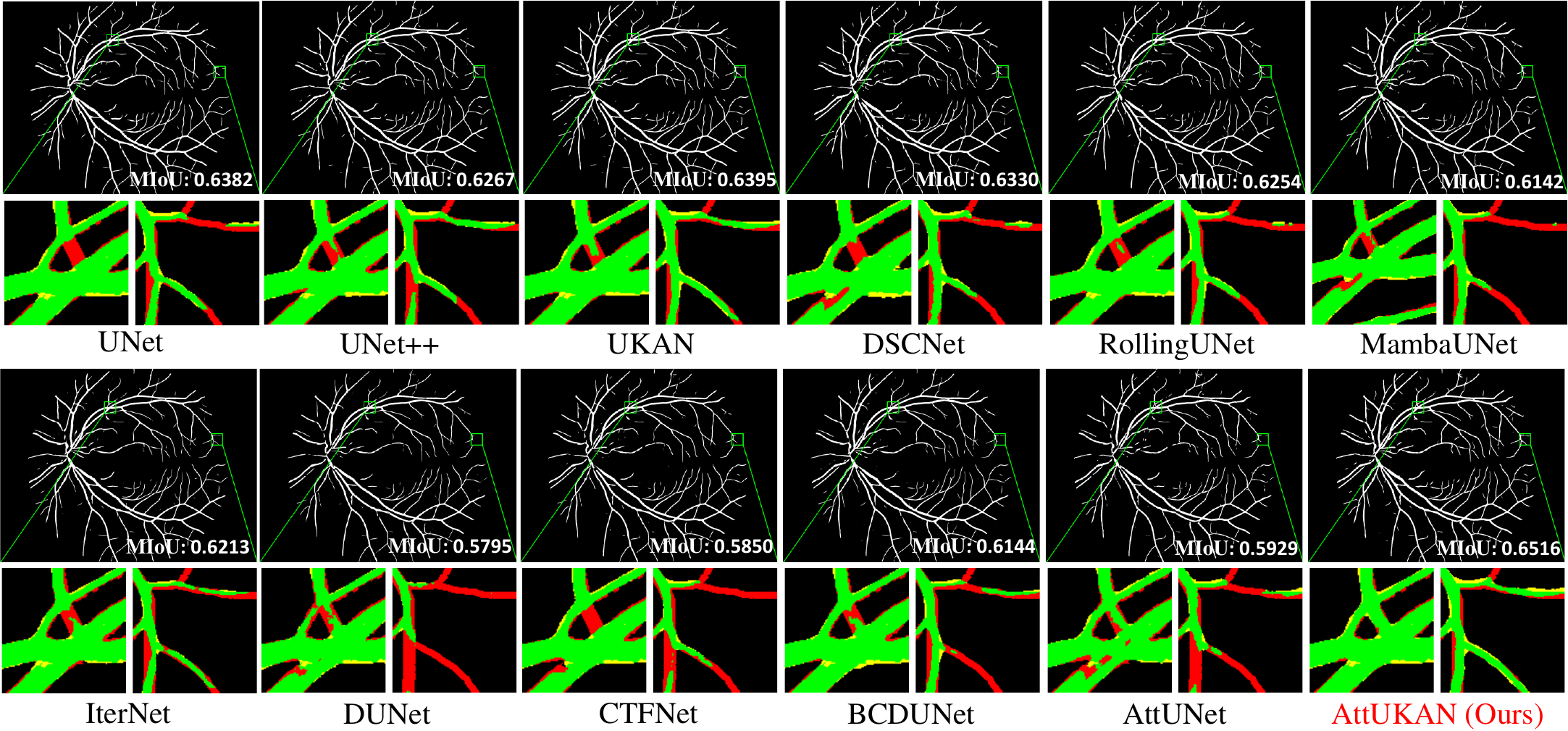}
\caption{Visualization of tiny vessel segmentation results on our private dataset. Green pixels represent true positive predictions; yellow pixels indicate false positive predictions; and red pixels denote false negatives predictions. Our proposed AttUKAN can get a more accurate segmentation result with more green regions and higher MIoU.}
\label{fig8}
\end{figure*}

\noindent{\textbf{Qualitative Results:}}
Additionally, the visualization of tiny vessel segmentation results compared with our proposed AttUKAN and other state-of-the-art methods on the DRIVE, STARE, CHASE$\_$DB1, HRF and our private datasets are also shown in Figures \ref{fig4}-\ref{fig8}, respectively. Through the observation, it can be noted that our proposed method achieves superior performance with a higher MIoU and more precise predictions of small vessels compared to the other 11 methods. This can be attributed to our designed Label-guided Pixel-wise Contrastive Loss, which helps in extracting more discriminative feature-level fine-grained representations.

% Table 7
\begin{table*}[t]
\centering
\scriptsize
\caption{The ablation studies for each component of our proposed method (best results shown in bold). The results show each of our proposed component can boost the segmentation performance.}\label{tab7}
\begin{tabular}{lcc|ccc|ccc}
\hline
\multicolumn{3}{c|}{\text { Settings }} & \multicolumn{3}{c|}{\text { DRIVE }} & \multicolumn{3}{c}{\text { STARE }}   \\
Framework & \itshape{AGs} & $\mathcal{L}_{LPCL}$ & $\text {F1}\uparrow$  & $\text {MIoU (\%)}\uparrow$ &$\text {AUC (\%)}\uparrow$ & $\text {F1 (\%)}\uparrow$ & $\text {MIoU (\%)}\uparrow$ & $\text {AUC (\%)}\uparrow$  \\
\hline
UKAN & & &$82.05$ &$69.58$ &$97.81$ &$80.39$ &$67.56$ &$97.42$ \\
AttUKAN (w/o LPCL) & \checkmark & &$82.28$ &$69.93$&$97.88$ &$80.88$ &$68.27$ &$97.62$ \\
UKAN (w LPCL) & & \checkmark &$82.30$ &$69.94$ &$97.86$ &$80.48$ &$67.67$ &$97.79$  \\
AttUKAN (Ours) & \checkmark & \checkmark &$\mathbf{82.50}$ &$\mathbf{70.24}$ &$\mathbf{97.95}$ &$\mathbf{81.14}$ &$\mathbf{68.64}$ &$\mathbf{97.96}$ \\
\hline
\end{tabular}
    
    % \bigskip
    
\begin{tabular}{lcc|ccc|ccc}
\hline
\multicolumn{3}{c|}{\text { Settings }} & \multicolumn{3}{c|}{\text { CHASE\text{\_}DB1 }} &\multicolumn{3}{c}{\text { HRF }} \\
Framework & \itshape{AGs} & $\mathcal{L}_{LPCL}$ & $\text {F1 (\%)}\uparrow$  & $\text {MIoU (\%)}\uparrow$ &$\text {AUC (\%)}\uparrow$ & $\text {F1 (\%)}\uparrow$  & $\text {MIoU (\%)}\uparrow$ &$\text {AUC (\%)}\uparrow$  \\
\hline
UKAN & & &$80.66$ &$67.63$ &$97.82$ &$80.11$ &$67.07$ &$97.51$ \\
AttUKAN (w/o LPCL) & \checkmark & &$81.25$ &$68.47$ &$98.13$ &$80.13$ &$67.08$ &$97.69$ \\
UKAN (w LPCL) & & \checkmark &$81.27$ &$68.49$ &$97.92$ &$80.16$ &$67.12$ &$97.46$ \\
AttUKAN (Ours) & \checkmark & \checkmark &$\mathbf{81.34}$ &$\mathbf{68.59}$ &$\mathbf{98.21}$ &$\mathbf{80.21}$ &$\mathbf{67.21}$ &$\mathbf{97.89}$ \\
\hline
\end{tabular}
% \bigskip

\begin{tabular}{lcc|ccc}
\hline
    \multicolumn{3}{c|}{\text { Settings }} &\multicolumn{3}{c}{\text { Our Private Dataset }} \\
    Framework & \itshape{AGs} & $\mathcal{L}_{LPCL}$ & $\text {F1 (\%)}\uparrow$  & $\text {MIoU (\%)}\uparrow$ &$\text {AUC (\%)}\uparrow$ \\
    \hline
    UKAN & & & $79.87$ &$66.65$&$97.43$\\
AttUKAN (w/o LPCL) & \checkmark &  & $80.08$ &$66.93$ &$97.69$ \\
    UKAN (w LPCL) & & \checkmark  & $79.93$ &$66.74$ &$97.41$ \\
    AttUKAN (Ours) & \checkmark & \checkmark & $\mathbf{80.09}$ &$\mathbf{66.94}$ &$\mathbf{97.73}$ \\
    \hline
\end{tabular}
\label{Ablation}
\end{table*}

% Table 8
\begin{table*}[ht]
\centering
\setlength{\tabcolsep}{2pt}
\caption{The ablation studies for each loss of our proposed method (best results shown in bold).}\label{tab8}
\scriptsize
\begin{tabular}{lcccc|ccc|ccc}
\hline
\multicolumn{5}{c|}{\text { Settings }} & \multicolumn{3}{c|}{\text { DRIVE }} & \multicolumn{3}{c}{\text { STARE }}   \\
Framework & $\mathcal{L}_{BCE}$ & $\mathcal{L}_{dice}$ & $\mathcal{L}_{jaccard}$ & $\mathcal{L}_{LPCL}$ & $\text {F1 (\%)}\uparrow$  & $\text {MIoU (\%)}\uparrow$ &$\text {AUC (\%)}\uparrow$ & $\text {F1 (\%)}\uparrow$ & $\text {MIoU (\%)}\uparrow$ & $\text {AUC (\%)}\uparrow$  \\
\hline
AttUKAN (BCE loss) & \checkmark  & & & &$81.73$ &$69.16$ &$88.35$ &$79.62$ &$66.61$ &$85.63$ \\
AttUKAN (BCE + dice loss) & \checkmark  & \checkmark & & &$82.16$ &$69.74$ &$90.34$ &$80.84$ &$68.20$ &$87.79$ \\
AttUKAN (w/o LPCL) & \checkmark & \checkmark & \checkmark & &$82.28$ &$69.93$&$97.88$ &$80.88$ &$68.27$ &$97.62$ \\
AttUKAN (Ours)  & \checkmark & \checkmark & \checkmark & \checkmark &$\mathbf{82.50}$ &$\mathbf{70.24}$ &$\mathbf{97.95}$ &$\mathbf{81.14}$ &$\mathbf{68.64}$ &$\mathbf{97.96}$ \\
\end{tabular}

% \bigskip

\begin{tabular}{lcccc|ccc|ccc}
\hline
\multicolumn{5}{c|}{\text { Settings }} & \multicolumn{3}{c|}{\text { CHASE\text{\_}DB1 }} & \multicolumn{3}{c}{\text { HRF }}   \\
Framework & $\mathcal{L}_{BCE}$ & $\mathcal{L}_{dice}$ & $\mathcal{L}_{jaccard}$ & $\mathcal{L}_{LPCL}$ & $\text {F1 (\%)}\uparrow$  & $\text {MIoU (\%)}\uparrow$ &$\text {AUC (\%)}\uparrow$ & $\text {F1 (\%)}\uparrow$ & $\text {MIoU (\%)}\uparrow$ & $\text {AUC (\%)}\uparrow$  \\
\hline
AttUKAN (BCE loss)& \checkmark  & & & &$81.02$ &$68.14$ &$89.18$ &$78.98$ &$65.54$ &$86.00$ \\
AttUKAN (BCE + dice loss) & \checkmark  & \checkmark & & &$81.22$ &$68.42$ &$89.29$ &$80.07$ &$67.00$ &$89.12$ \\
AttUKAN (w/o LPCL) & \checkmark & \checkmark & \checkmark & &$81.25$ &$68.47$ &$98.13$ &$80.13$ &$67.08$ &$97.69$ \\
AttUKAN (Ours)  & \checkmark & \checkmark & \checkmark & \checkmark &$\mathbf{81.34}$ &$\mathbf{68.59}$ &$\mathbf{98.21}$ &$\mathbf{80.21}$ &$\mathbf{67.21}$ &$\mathbf{97.89}$ \\
\hline
\end{tabular}
% \bigskip

\begin{tabular}{lcccc|ccc}
% \hline
 \multicolumn{5}{c|}{\text { Settings }} &\multicolumn{3}{c}{\text { Our Private Dataset }} \\
Framework & $\mathcal{L}_{BCE}$ & $\mathcal{L}_{dice}$ & $\mathcal{L}_{jaccard}$ & $\mathcal{L}_{LPCL}$ & $\text {F1 (\%)}\uparrow$  & $\text {MIoU (\%)}\uparrow$ &$\text {AUC (\%)}\uparrow$ \\
 \hline
 AttUKAN (BCE loss) & \checkmark  & & & & $79.74$ &$66.51$ &$\mathbf{97.86}$\\
 AttUKAN (BCE + dice loss)& \checkmark  & \checkmark & & &$79.83$ &$66.59$ &$97.44$ \\
 AttUKAN (w/o LPCL)  & \checkmark & \checkmark & \checkmark & &$80.08$ &$66.93$ &$97.69$ \\
 AttUKAN (Ours)  & \checkmark & \checkmark & \checkmark & \checkmark &$\mathbf{80.09}$ &$\mathbf{66.94}$ &$97.73$ \\
 \hline
\end{tabular}
\label{Ablation}
\end{table*}

% Table 9

%ablation lamda
\begin{table*}[ht]
\centering
\scriptsize
\caption{The ablation studies of weighting coefficient $\lambda_4$ of LPCL. (best results shown in bold).}\label{tab9}
\begin{tabular}{l|ccc|ccc|ccc}
\hline
\multirow{2}{*}{$\lambda_4$} & \multicolumn{3}{c|}{\text { DRIVE }} & \multicolumn{3}{c|}{\text { STARE }} & \multicolumn{3}{c}{\text { CHASE\text{\_}DB1 }} \\
& $\text {F1 (\%)}\uparrow$  & $\text {MIoU (\%)}\uparrow$ & $\text {AUC (\%)}\uparrow$ & $\text {F1 (\%)}\uparrow$  & $\text {MIoU (\%)}\uparrow$ & $\text {AUC (\%)}\uparrow$ & $\text {F1 (\%)}\uparrow$  & $\text {MIoU (\%)}\uparrow$ & $\text {AUC (\%)}\uparrow$  \\
\hline
0.0 &$82.28$ &$69.93$ &$97.88$ &$80.88$ &$68.27$ &$97.62$ &$81.25$ &$68.47$ &$98.13$\\
0.1 &$82.19$ &$69.80$ &$97.84$ &$79.96$ &$66.98$ &$97.12$ &$81.05$ &$68.17$  &$98.19$ \\
0.3 &$\mathbf{82.50}$ &$\mathbf{70.24}$ &$\mathbf{97.95}$ &$\mathbf{81.14}$ &$\mathbf{68.64}$ &$\mathbf{97.96}$ &$\mathbf{81.34}$ &$\mathbf{68.59}$ &$\mathbf{98.21}$ \\
0.5 &$82.10$ &$69.66$ &$97.86$ &$80.12$ &$67.22$ &$97.36$ &$81.17$ &$68.35$ &$98.17$  \\
0.7 &$82.20$ &$69.82$ &$97.90$ &$80.58$ &$67.82$ &$97.85$ &$81.06$ &$68.20$ &$98.07$ \\
0.9 &$82.16$ &$69.75$ &$97.86$ &$80.92$ &$68.31$ &$97.90$ &$80.85$ &$67.90$ &$98.06$ \\
1.0 &$82.22$ &$69.83$ &$97.88$ &$80.98$ &$68.41$ &$97.77$ &$81.03$ &$68.15$ &$98.16$ \\
\hline
\end{tabular}
% \bigskip
\begin{tabular}{l|ccc|ccc}

\multirow{2}{*}{$\lambda_4$}  &\multicolumn{3}{c|}{\text { HRF }} & \multicolumn{3}{c}{\text { Our Private Dataset }}  \\
& $\text {F1 (\%)}\uparrow$  & $\text {MIoU (\%)}\uparrow$ & $\text {AUC (\%)}\uparrow$ & $\text {F1 (\%)}\uparrow$  & $\text {MIoU (\%)}\uparrow$ &$\text {AUC (\%)}\uparrow$  \\
\hline
0.0 &$80.13$ &$67.08$ &$97.69$ &$80.08$ &$66.93$ &$97.69$ \\
0.1 &$79.75$ &$66.58$ &$97.53$ &$79.77$ &$66.52$ &$97.57$ \\
0.3 &$79.96$ &$66.86$ &$97.56$ &$79.37$ &$65.97$ &$97.48$ \\
0.5 &$\mathbf{80.21}$ &$\mathbf{67.21}$&$\mathbf{97.89}$ &$\mathbf{80.09}$ &$\mathbf{66.94}$ &$\mathbf{97.73}$\\
0.7 &$80.21$ &$67.21$ &$97.78$ &$79.86$ &$66.63$ &$97.63$ \\
0.9 &$79.80$ &$66.63$ &$97.59$ &$79.62$ &$66.31$ &$97.22$ \\
1.0 &$80.12$ &$67.09$ &$97.62$ &$79.91$ &$66.71$ &$97.48$\\
\hline
\end{tabular}
\label{Ablation}
\end{table*}

% Table 10
\begin{table*}[t]
\centering
\scriptsize
\setlength{\tabcolsep}{1.5pt}
\caption{The ablation studies of LPCL with different level features. The results show that our AttUKAN can extract more discriminative fine-grained feature-level representations using higher-level features ($5_{th}$-level)}\label{tab10}
\begin{tabular}{l|ccc|ccc|ccc}
\hline
\multirow{2}{*}{Features} & \multicolumn{3}{c|}{\text { DRIVE }} & \multicolumn{3}{c|}{\text { STARE }} & \multicolumn{3}{c}{\text { CHASE\_DB1 }} \\
& $\text {F1 (\%)}\uparrow$  & $\text {MIoU (\%)}\uparrow$ & $\text {AUC (\%)}\uparrow$ & $\text {F1 (\%)}\uparrow$  & $\text {MIoU (\%)}\uparrow$ & $\text {AUC (\%)}\uparrow$ & $\text {F1 (\%)}\uparrow$  & $\text {MIoU (\%)}\uparrow$ & $\text {AUC (\%)}\uparrow$  \\
\hline
$3_{th}$-level feature &$82.23$ &$69.87$ &$89.63$ &$81.04$ &$88.25$ &$88.25$ &$81.08$ &$68.23$ &$89.25$ \\
$4_{th}$-level feature &$81.96$ &$69.44$ &$91.14$ &$80.23$ &$67.43$ &$87.35$ &$81.16$ &$68.34$ &$89.39$ \\
$5_{th}$-level feature (bottleneck) &$\mathbf{82.50}$ &$\mathbf{70.24}$ &$\mathbf{97.95}$ &$\mathbf{81.14}$ &$\mathbf{68.64}$ &$\mathbf{97.96}$ &$\mathbf{81.34}$ &$\mathbf{68.59}$ &$\mathbf{98.21}$ \\
\hline
\end{tabular}

%\bigskip

\begin{tabular}{l|ccc|ccc}
\multirow{2}{*}{Features}  &\multicolumn{3}{c|}{\text { HRF }} & \multicolumn{3}{c}{\text { Our Private Dataset }}  \\
& $\text {F1 (\%)}\uparrow$  & $\text {MIoU (\%)}\uparrow$ & $\text {AUC (\%)}\uparrow$ & $\text {F1 (\%)}\uparrow$  & $\text {MIoU (\%)}\uparrow$ &$\text {AUC (\%)}\uparrow$  \\
\hline
$3_{th}$-level feature &$80.00$ &$66.92$ &$88.37$ &$77.60$ &$63.61$ &$96.29$ \\
$4_{th}$-level feature &$80.05$ &$66.97$ &$88.90$ &$78.12$ &$64.32$ &$96.96$ \\
$5_{th}$-level feature (bottleneck) &$\mathbf{80.21}$ &$\mathbf{67.21}$ &$\mathbf{97.89}$ &$\mathbf{80.09}$ &$\mathbf{66.94}$ &$\mathbf{97.73}$ \\
\hline
\end{tabular}
\label{Ablation}
\end{table*}

% Table 11
%ablation of LPCL
\begin{table*}[h]
\centering
\setlength{\tabcolsep}{2pt}
\scriptsize
\caption{The ablation studies of LPCL on different networks. The results show our proposed LCPL can incorporate into different networks, verifying its generalization ability.}\label{tab11}
\begin{tabular}{l|ccc|ccc|ccc}
\Xhline{1.0pt}
\multirow{2}{*}{Methods} & \multicolumn{3}{c|}{\text { DRIVE }} & \multicolumn{3}{c|}{\text { STARE }} & \multicolumn{3}{c}{\text { CHASE\text{\_}DB1 }} \\
& $\text {F1 (\%)}\uparrow$  & $\text {MIoU (\%)}\uparrow$ & $\text {AUC (\%)}\uparrow$ & $\text {F1 (\%)}\uparrow$  & $\text {MIoU (\%)}\uparrow$ & $\text {AUC (\%)}\uparrow$ & $\text {F1 (\%)}\uparrow$  & $\text {MIoU (\%)}\uparrow$ & $\text {AUC (\%)}\uparrow$  \\
\Xhline{1.0pt}
AttUNet (w/o LPCL) & $80.48$ & $67.39$ & $87.42$ & $79.82$ & $66.79$ & $85.92$ & $79.94$ & $66.64$ & $87.32$ \\
AttUNet (w LPCL) & $81.03$ & $68.14$ & $88.79$ & $80.15$ & $67.25$ & $86.34$ & $80.33$ & $67.17$ & $87.75$ \\

BCDUNet (w/o LPCL) & $82.00$ & $69.51$ & $90.68$ & $80.24$ & $67.44$ & $88.67$ & $79.12$ & $65.51$ & $87.88$ \\
BCDUNet (w LPCL) & $82.23$ & $69.86$ & $89.47$ & $80.90$ & $68.36$ & $87.98$ & $79.63$ & $66.22$ & $88.77$ \\

DUNet (w/o LPCL) & $81.69$ & $69.09$ & $89.25$ & $77.71$ & $64.10$ & $84.59$ & $79.30$ & $65.75$ & $88.36$ \\
DUNet (w LPCL) & $81.85$ & $69.32$ & $89.31$ & $78.56$ & $65.23$ & $85.69$ & $80.19$ & $66.99$ & $89.27$ \\

RollingUNet (w/o LPCL) & $82.00$ & $69.53$ & $89.61$ & $79.86$ & $66.93$ & $87.00$ & $80.41$ & $67.30$ & $88.59$ \\
RollingUNet (w LPCL) & $82.28$ & $69.92$ & $90.17$ & $81.13$ & $68.61$ & $88.71$ & $80.70$ & $67.70$ & $88.90$ \\

DSCNet (w/o LPCL) & $80.89$ & $67.95$ & $88.96$ & $78.24$ & $64.58$ & $85.35$ & $78.08$ & $64.09$ & $87.57$ \\
DSCNet (w LPCL) & $81.25$ & $68.47$ & $88.93$ & $79.24$ & $65.91$ & $87.26$ & $78.25$ & $64.33$ & $87.63$ \\

UKAN (w/o LPCL) &$82.05$ &$69.58$ &$97.81$ &$80.39$ &$67.56$ &$97.42$&$80.66$&$67.63$&$97.82$ \\
UKAN (w LPCL) & $82.30$ & $69.94$&$97.86$ & $80.48$&$67.67$ & $97.79$ & $81.27$&$68.49$&$97.92$ \\
\Xhline{1.0pt}
\end{tabular}
% \bigskip
\begin{tabular}{l|ccc|ccc}
\multirow{2}{*}{Methods}  &\multicolumn{3}{c|}{\text { HRF }} & \multicolumn{3}{c}{\text { Our Private Dataset }}  \\
& $\text {F1 (\%)}\uparrow$  & $\text {MIoU (\%)}\uparrow$ & $\text {AUC (\%)}\uparrow$ & $\text {F1 (\%)}\uparrow$  & $\text {MIoU (\%)}\uparrow$ &$\text {AUC (\%)}\uparrow$  \\
\Xhline{1.0pt}
AttUNet (w/o LPCL) & $78.60$ & $65.04$ & $86.89$ & $76.97$ & $62.77$ & $96.03$ \\
AttUNet (w LPCL) & $79.09$ & $65.68$ & $87.45$ & $78.07$ & $64.26$ & $97.17$ \\
BCDUNet (w/o LPCL) & $77.46$ & $63.49$ & $86.65$ & $76.27$ & $61.86$ & $95.27$ \\
BCDUNet (w LPCL) & $77.56$ & $63.64$ & $87.19$ & $76.33$ & $61.96$ & $95.82$ \\
DUNet (w/o LPCL) & $77.93$ & $64.11$ & $86.79$ & $72.88$ & $57.58$ & $93.43$ \\
DUNet (w LPCL) & $77.97$ & $64.18$ & $86.53$ & $75.37$ & $60.71$ & $93.99$ \\
RollingUNet (w/o LPCL) & $79.42$ & $66.13$ & $88.33$ & $73.85$ & $58.75$ & $92.90$ \\
RollingUNet (w LPCL) & $79.78$ & $66.60$ & $89.04$ & $75.59$ & $61.02$ & $95.01$ \\
DSCNet (w/o LPCL) & $75.64$ & $61.20$ & $85.16$ & $76.15$ & $61.70$ & $96.09$ \\
DSCNet (w LPCL) & $76.32$ & $62.08$ & $86.32$ & $76.28$ & $61.86$ & $96.10$ \\
UKAN (w/o LPCL) & $80.11$ & $67.07$ & $97.51$ & $79.87$& $66.65$&$97.43$ \\
UKAN (w LPCL) &$80.16$&$67.12$&$97.46$&$79.93$&$66.74$&$97.41$ \\
\Xhline{1.0pt}
\end{tabular}
\label{Ablation}
\end{table*}

\section{Discussion}\label{sec5}

\subsection{Ablation studies for the components of AttUKAN}
Ablation studies have been carried out to verify the effectiveness of each part of our proposed components, including the Label-guided Pixel-wise Contrastive Loss  ($\mathcal{L}_{LPCL}$) and the implementation of Attention Gates ($AGs$). UKAN is used as the baseline and we apply each component to UKAN, respectively. In order for fair comparison, we also use binary cross-entropy loss ($\mathcal{L}_{BCE}$), dice loss ($\mathcal{L}_{dice}$) and jaccard loss ($\mathcal{L}_{jaccard}$) to optimize the baseline network UKAN.
Finally, the results from the ablation experiments on the DRIVE, STARE, CHASE$\_$DB1, HRF and our private dataset are shown in Table \ref{tab7}. It can be observed that each part of our proposed components contributes to the improvement of performance. Specifically, the comparison shows that the introduction of $\mathcal{L}_{LPCL}$ and $AGs$ improves the performance of our AttUKAN framework, with a promotion in the F1, MIoU and AUC metrics. As shown in Table \ref{tab7}, the components have improved the F1 score by 0.45\%, 0.75\%, 0.68\%, 0.10\% and 0.22\% on the DRIVE, STARE, CHASE\_DB1, HRF and our private datasets, respectively, along with improvements in the MIoU scores by 0.66\%, 1.08\%, 0.96\%, 0.14\% and 0.29\%.

%ablation components
\subsection{Ablation studies for each loss of AttUKAN}

Ablation studies have also been carried out to verify every baseline loss, including binary cross-entropy loss ($\mathcal{L}_{BCE}$), dice loss ($\mathcal{L}_{dice}$), jaccard loss ($\mathcal{L}_{jaccard}$) and our proposed Label-guided Pixel-wise Contrastive Loss  ($\mathcal{L}_{LPCL}$). It can be observed from Table \ref{tab8} that each loss contributes to the improvement of performance. And finally our proposed AttUKAN with $\mathcal{L}_{BCE}$, $\mathcal{L}_{dice}$, $\mathcal{L}_{jaccard}$ and $\mathcal{L}_{LPCL}$ achieves the highest performance among the 5 retinal vessel datasets. 

\subsection{Ablation study of $\lambda_4$}
Additionally, we also conduct an ablation study on the Label-guided Pixel-wise Contrastive Loss (LPCL), to determine the weighting coefficient $\lambda_4$ of LPCL in our final hybrid loss. As shown in the Table \ref{tab9}, we conclude that when $\lambda_4=0.3$, the segmentation performance of the model is the most optimal on DRIVE, STARE and CHASE\_DB1 datasets and $\lambda_4=0.5$ is the best setting on the HRF and our private dataset. 

\subsection{Ablation study of LPCL with different level features}
In order to thoroughly evaluate the capabilities of LPCL in extracting discriminative representations, we conduct an ablation study using LPCL with different level features. According to Section \ref{sec3.4}, we optimize LPCL by forwarding feature maps from different layers in AttUKAN. Specifically, the $3_{rd}$-level feature ${\boldsymbol{X}}^3$, $4_{th}$-level feature ${\boldsymbol{X}}^4$, and $5_{th}$-level feature ${\boldsymbol{X}}^5$ are implemented in LPCL in this study, and results on the five datasets are shown in Table \ref{tab10}. The results indicate that LPCL with the feature forwarded from the bottleneck of the model ($5_{th}$-level feature) shows the best performance across all datasets, demonstrating that our AttUKAN can extract more discriminative fine-grained feature-level representations using higher-level features.

\subsection{Ablation study of LPCL on different networks}
To further explore the effectiveness of LPCL, we carry out an ablation study of only LPCL with other baseline methods. To ensure the fairness and objectivity of the experiments, we first configure the same combination of binary cross-entropy loss ($\mathcal{L}_{BCE}$), dice loss ($\mathcal{L}_{dice}$) and jaccard loss ($\mathcal{L}_{jaccard}$) on the selected networks, including AttUNet \cite{oktay2018attention}, BCDUNet \cite{azad2019bi}, DUNet \cite{jin2019dunet}, RollingUNet \cite{liu2024rolling}, and DSCNet \cite{qi2023dynamic}. LPCL is then applieded to each network respectively for a comprehensive comparison of the different models under consistent conditions. The results from the ablation experiments on the DRIVE, STARE, CHASE\_DB1, HRF, and our private dataset are presented in Table \ref{tab11}, indicating performance enhancements across each network through improvements in F1 scrore, MIOU, and AUC metrics.
Notably, for the quantitative results shown from Table \ref{tab1}-\ref{tab5}, we follow its original loss used in each baseline method, while on Table \ref{tab11}, all the baseline networks are under the same loss setting (w/o LPCL vs. w LPCL).

\section{Conclusion}\label{sec6}

In this work, a novel retinal vessel segmentation network named AttUKAN is proposed. Specifically, we incorporate Attention Gates into UKAN to selectively filter the features passed through skip connections, thereby enhancing the performance of the model across various datasets. Additionally, we design a novel contrastive loss function named Label-guided Pixel-wise Contrastive Loss to supervise the extraction of more discriminative fine-grained feature-level representations. Experiments on four public datasets and our private dataset are conducted to evaluate our proposed AttUKAN. The experimental results demonstrate that our AttUKAN can achieve outstanding performance compared to existing state-of-the-art retinal vessel segmentation methods. 

\section*{Acknowledgments}
This work was supported in part by the Natural Science Foundation of China under Grant 82371112, 623B2001, 62394311, in part by Beijing Natural Science Foundation under Grant Z210008, and in part by High-grade, Precision and Advanced University Discipline Construction Project of Beijing (BMU2024GJJXK004).   

% \section*{Author contributions}

% %delete if not applicable
% [Examples: X.Y. was involved in conceptualization, investigation, writing—original draft, project management. A.B. and Y.Z. were involved in investigation, writing—review and editing.] 

% \section*{Financial disclosure}

% [Example: None reported.]


% The authors declare no potential conflict of interests.

% \section*{Data Availability Statement}
% %Required. See Wiley’s Data Sharing Policies for examples (https://authorservices.wiley.com/author-resources/Journal-Authors/open-access/data-sharing-citation/data-sharing-policy.html). \\
% [Examples:\\Statement for Data available on request from the authors: "The data that support the findings of this study are available from the corresponding author upon reasonable request.”\\
% If no data are being shared, just state “Research data are not shared.”]

% \section*{Supporting information}

% %delete if not applicable
% The following supporting information is available as part of the online article:

% \noindent
% \textbf{Figure S1.}
% [Legend]

% \noindent
% \textbf{Figure S2.}
% [Legend]

%Please submit any supporting information as a separate file. It will be published as Supporting Information online with free access.

\nocite{*}% Show all bib entries - both cited and uncited; comment this line to view only cited bib entries;
\bibliography{wileyNJD-ACS}%

\providecommand{\url}[1]{\texttt{#1}}
\providecommand{\urlprefix}{}
\providecommand{\foreignlanguage}[2]{#2}
\providecommand{\Capitalize}[1]{\uppercase{#1}}
\providecommand{\capitalize}[1]{\expandafter\Capitalize#1}
\providecommand{\bibliographycite}[1]{\cite{#1}}
\providecommand{\bbland}{and}
\providecommand{\bblchap}{chap.}
\providecommand{\bblchapter}{chapter}
\providecommand{\bbletal}{et~al.}
\providecommand{\bbleditors}{editors}
\providecommand{\bbleds}{eds: }
\providecommand{\bbleditor}{editor}
\providecommand{\bbled}{ed.}
\providecommand{\bbledition}{edition}
\providecommand{\bbledn}{ed.}
\providecommand{\bbleidp}{page}
\providecommand{\bbleidpp}{pages}
\providecommand{\bblerratum}{erratum}
\providecommand{\bblin}{in}
\providecommand{\bblmthesis}{Master's thesis}
\providecommand{\bblno}{no.}
\providecommand{\bblnumber}{number}
\providecommand{\bblof}{of}
\providecommand{\bblpage}{page}
\providecommand{\bblpages}{pages}
\providecommand{\bblp}{p}
\providecommand{\bblphdthesis}{Ph.D. thesis}
\providecommand{\bblpp}{pp}
\providecommand{\bbltechrep}{}
\providecommand{\bbltechreport}{Technical Report}
\providecommand{\bblvolume}{volume}
\providecommand{\bblvol}{Vol.}
\providecommand{\bbljan}{January}
\providecommand{\bblfeb}{February}
\providecommand{\bblmar}{March}
\providecommand{\bblapr}{April}
\providecommand{\bblmay}{May}
\providecommand{\bbljun}{June}
\providecommand{\bbljul}{July}
\providecommand{\bblaug}{August}
\providecommand{\bblsep}{September}
\providecommand{\bbloct}{October}
\providecommand{\bblnov}{November}
\providecommand{\bbldec}{December}
\providecommand{\bblfirst}{First}
\providecommand{\bblfirsto}{1st}
\providecommand{\bblsecond}{Second}
\providecommand{\bblsecondo}{2nd}
\providecommand{\bblthird}{Third}
\providecommand{\bblthirdo}{3rd}
\providecommand{\bblfourth}{Fourth}
\providecommand{\bblfourtho}{4th}
\providecommand{\bblfifth}{Fifth}
\providecommand{\bblfiftho}{5th}
\providecommand{\bblst}{st}
\providecommand{\bblnd}{nd}
\providecommand{\bblrd}{rd}
\providecommand{\bblth}{th}
\begin{thebibliography}{10}

\bibitem{smart2015study}
Thomas~J Smart, Christopher~J Richards, Rhythm Bhatnagar, Carlos Pavesio,
  Rupesh Agrawal, Philip~H Jones, \bblin{} {\it Optical trapping and optical
  micromanipulation XII}, SPIE, \textbf{2015}, \bblpp{}. 342--348.

\bibitem{ding2014retinal}
Jie Ding, Khin~Lay Wai, Kevin McGeechan, M~Kamran Ikram, Ryo Kawasaki, Jing
  Xie, Ronald Klein, Barbara~BK Klein, Mary~Frances Cotch, Jie~Jin Wang,
  \bbletal{}, {\it Journal of hypertension} \textbf{2014}, {\it 32} (2),
  207--215.

\bibitem{na2018retinal}
Tong Na, Jianyang Xie, Yitian Zhao, Yifan Zhao, Yue Liu, Yongtian Wang, Jiang
  Liu, {\it Medical physics} \textbf{2018}, {\it 45} (7), 3132--3146.

\bibitem{han2014blood}
Zhe Han, Yilong Yin, Xianjing Meng, Gongping Yang, Xiaowei Yan, \bblin{} {\it
  2014 IEEE International Conference on Data Mining Workshop}, IEEE,
  \textbf{2014}, \bblpp{}. 960--967.

\bibitem{long2015fully}
Jonathan Long, Evan Shelhamer, Trevor Darrell, \bblin{} {\it Proceedings of the
  IEEE conference on computer vision and pattern recognition}, \textbf{2015},
  \bblpp{}. 3431--3440.

\bibitem{ronneberger2015u}
Olaf Ronneberger, Philipp Fischer, Thomas Brox, \bblin{} {\it Medical image
  computing and computer-assisted intervention--MICCAI 2015: 18th international
  conference, Munich, Germany, October 5-9, 2015, proceedings, part III 18},
  Springer, \textbf{2015}, \bblpp{}. 234--241.

\bibitem{jin2019dunet}
Qiangguo Jin, Zhaopeng Meng, Tuan~D Pham, Qi~Chen, Leyi Wei, Ran Su, {\it
  Knowledge-Based Systems} \textbf{2019}, {\it 178}, 149--162.

\bibitem{qi2023dynamic}
Yaolei Qi, Yuting He, Xiaoming Qi, Yuan Zhang, Guanyu Yang, \bblin{} {\it
  Proceedings of the IEEE/CVF International Conference on Computer Vision},
  \textbf{2023}, \bblpp{}. 6070--6079.

\bibitem{li2020iternet}
Liangzhi Li, Manisha Verma, Yuta Nakashima, Hajime Nagahara, Ryo Kawasaki,
  \bblin{} {\it Proceedings of the IEEE/CVF winter conference on applications
  of computer vision}, \textbf{2020}, \bblpp{}. 3656--3665.

\bibitem{wang2020ctf}
Kun Wang, Xiaohong Zhang, Sheng Huang, Qiuli Wang, Feiyu Chen, \bblin{} {\it
  2020 IEEE 17th International Symposium on Biomedical Imaging (ISBI)}, IEEE,
  \textbf{2020}, \bblpp{}. 1237--1241.

\bibitem{azad2019bi}
Reza Azad, Maryam Asadi-Aghbolaghi, Mahmood Fathy, Sergio Escalera, \bblin{}
  {\it Proceedings of the IEEE/CVF international conference on computer vision
  workshops}, \textbf{2019}, \bblpp{}. 0--0.

\bibitem{oktay2018attention}
Ozan Oktay, Jo~Schlemper, Loic~Le Folgoc, Matthew Lee, Mattias Heinrich,
  Kazunari Misawa, Kensaku Mori, Steven McDonagh, Nils~Y Hammerla, Bernhard
  Kainz, \bbletal{}, {\it arXiv preprint arXiv:1804.03999} \textbf{2018}.

\bibitem{zhou2018unet++}
Zongwei Zhou, Md~Mahfuzur Rahman~Siddiquee, Nima Tajbakhsh, Jianming Liang,
  \bblin{} {\it Deep Learning in Medical Image Analysis and Multimodal Learning
  for Clinical Decision Support: 4th International Workshop, DLMIA 2018, and
  8th International Workshop, ML-CDS 2018, Held in Conjunction with MICCAI
  2018, Granada, Spain, September 20, 2018, Proceedings 4}, Springer,
  \textbf{2018}, \bblpp{}. 3--11.

\bibitem{liu2024rolling}
Yutong Liu, Haijiang Zhu, Mengting Liu, Huaiyuan Yu, Zihan Chen, Jie Gao,
  \bblin{} {\it Proceedings of the AAAI Conference on Artificial Intelligence},
  \textbf{2024}, \bblpp{}. 3819--3827.

\bibitem{wang2024mamba}
Ziyang Wang, Jian-Qing Zheng, Yichi Zhang, Ge~Cui, Lei Li, {\it arXiv preprint
  arXiv:2402.05079} \textbf{2024}.

\bibitem{liu2024kan}
Ziming Liu, Yixuan Wang, Sachin Vaidya, Fabian Ruehle, James Halverson, Marin
  Solja{\v{c}}i{\'c}, Thomas~Y Hou, Max Tegmark, {\it arXiv preprint
  arXiv:2404.19756} \textbf{2024}.

\bibitem{li2024u}
Chenxin Li, Xinyu Liu, Wuyang Li, Cheng Wang, Hengyu Liu, Yixuan Yuan, {\it
  arXiv preprint arXiv:2406.02918} \textbf{2024}.

\bibitem{chen2020simple}
Ting Chen, Simon Kornblith, Mohammad Norouzi, Geoffrey Hinton, \bblin{} {\it
  International conference on machine learning}, PMLR, \textbf{2020}, \bblpp{}.
  1597--1607.

\bibitem{he2020momentum}
Kaiming He, Haoqi Fan, Yuxin Wu, Saining Xie, Ross Girshick, \bblin{} {\it
  Proceedings of the IEEE/CVF conference on computer vision and pattern
  recognition}, \textbf{2020}, \bblpp{}. 9729--9738.

\bibitem{singh2015local}
Nagendra~Pratap Singh, Rajesh Kumar, Rajeev Srivastava, \bblin{} {\it
  International Conference on Computing, Communication \& Automation}, IEEE,
  \textbf{2015}, \bblpp{}. 1166--1170.

\bibitem{zana2001segmentation}
Frederic Zana, J-C Klein, {\it IEEE transactions on image processing}
  \textbf{2001}, {\it 10} (7), 1010--1019.

\bibitem{oliveira2016unsupervised}
Wendeson~S Oliveira, Joyce~Vitor Teixeira, Tsang~Ing Ren, George~DC Cavalcanti,
  Jan Sijbers, {\it PloS one} \textbf{2016}, {\it 11} (2), e0149943.

\bibitem{fu2016deepvessel}
Huazhu Fu, Yanwu Xu, Stephen Lin, Damon~Wing Kee~Wong, Jiang Liu, \bblin{} {\it
  Medical Image Computing and Computer-Assisted Intervention--MICCAI 2016: 19th
  International Conference, Athens, Greece, October 17-21, 2016, Proceedings,
  Part II 19}, Springer, \textbf{2016}, \bblpp{}. 132--139.

\bibitem{guo2020bscn}
Yanfei Guo, Yanjun Peng, {\it BMC medical imaging} \textbf{2020}, {\it 20},
  1--22.

\bibitem{shin2019deep}
Seung~Yeon Shin, Soochahn Lee, Il~Dong Yun, Kyoung~Mu Lee, {\it Medical image
  analysis} \textbf{2019}, {\it 58}, 101556.

\bibitem{chen2024tokenunify}
Yinda Chen, Haoyuan Shi, Xiaoyu Liu, Te~Shi, Ruobing Zhang, Dong Liu, Zhiwei
  Xiong, Feng Wu, {\it arXiv preprint arXiv:2405.16847} \textbf{2024}.

\bibitem{dosovitskiy2020image}
Alexey Dosovitskiy, Lucas Beyer, Alexander Kolesnikov, Dirk Weissenborn,
  Xiaohua Zhai, Thomas Unterthiner, Mostafa Dehghani, Matthias Minderer, Georg
  Heigold, Sylvain Gelly, \bbletal{}, {\it arXiv preprint arXiv:2010.11929}
  \textbf{2020}.

\bibitem{cao2022conv}
Jinming Cao, Yangyan Li, Mingchao Sun, Ying Chen, Dani Lischinski, Daniel
  Cohen-Or, Baoquan Chen, Changhe Tu, {\it IEEE Transactions on Image
  Processing} \textbf{2022}, {\it 31}, 3726--3736.

\bibitem{ba2016layer}
Jimmy~Lei Ba, Jamie~Ryan Kiros, Geoffrey~E Hinton, {\it arXiv preprint
  arXiv:1607.06450} \textbf{2016}.

\bibitem{bahdanau2014neural}
Dzmitry Bahdanau, Kyunghyun Cho, Yoshua Bengio, {\it arXiv preprint
  arXiv:1409.0473} \textbf{2014}.

\bibitem{staal2004ridge}
Joes Staal, Michael~D Abr{\`a}moff, Meindert Niemeijer, Max~A Viergever, Bram
  Van~Ginneken, {\it IEEE transactions on medical imaging} \textbf{2004}, {\it
  23} (4), 501--509.

\bibitem{hoover2000locating}
AD~Hoover, Valentina Kouznetsova, Michael Goldbaum, {\it IEEE Transactions on
  Medical imaging} \textbf{2000}, {\it 19} (3), 203--210.

\bibitem{owen2009measuring}
Christopher~G Owen, Alicja~R Rudnicka, Robert Mullen, Sarah~A Barman, Dorothy
  Monekosso, Peter~H Whincup, Jeffrey Ng, Carl Paterson, {\it Investigative
  ophthalmology \& visual science} \textbf{2009}, {\it 50} (5), 2004--2010.

\bibitem{odstrcilik2013retinal}
Jan Odstrcilik, Radim Kolar, Attila Budai, Joachim Hornegger, Jiri Jan, Jiri
  Gazarek, Tomas Kubena, Pavel Cernosek, Ondrej Svoboda, Elli Angelopoulou,
  {\it IET Image Processing} \textbf{2013}, {\it 7} (4), 373--383.

\bibitem{gegundez2011function}
Manuel~Emilio Geg{\'u}ndez-Arias, Arturo Aquino, Jos{\'e}~Manuel Bravo, Diego
  Mar{\'\i}n, {\it IEEE transactions on medical imaging} \textbf{2011}, {\it
  31} (2), 231--239.

\bibitem{wang2019dual}
Bo~Wang, Shuang Qiu, Huiguang He, \bblin{} {\it Medical Image Computing and
  Computer Assisted Intervention--MICCAI 2019: 22nd International Conference,
  Shenzhen, China, October 13--17, 2019, Proceedings, Part I 22}, Springer,
  \textbf{2019}, \bblpp{}. 84--92.

\bibitem{anderson2018bottom}
Peter Anderson, Xiaodong He, Chris Buehler, Damien Teney, Mark Johnson, Stephen
  Gould, Lei Zhang, \bblin{} {\it Proceedings of the IEEE conference on
  computer vision and pattern recognition}, \textbf{2018}, \bblpp{}.
  6077--6086.

\bibitem{jetley2018learn}
Saumya Jetley, Nicholas~A Lord, Namhoon Lee, Philip~HS Torr, {\it arXiv
  preprint arXiv:1804.02391} \textbf{2018}.

\bibitem{hu2018squeeze}
Jie Hu, Li~Shen, Gang Sun, \bblin{} {\it Proceedings of the IEEE conference on
  computer vision and pattern recognition}, \textbf{2018}, \bblpp{}.
  7132--7141.

\bibitem{woo2018cbam}
Sanghyun Woo, Jongchan Park, Joon-Young Lee, In~So Kweon, \bblin{} {\it
  Proceedings of the European conference on computer vision (ECCV)},
  \textbf{2018}, \bblpp{}. 3--19.

\bibitem{soares2006retinal}
Jo{\~a}o~VB Soares, Jorge~JG Leandro, Roberto~M Cesar, Herbert~F Jelinek,
  Michael~J Cree, {\it IEEE Transactions on medical Imaging} \textbf{2006},
  {\it 25} (9), 1214--1222.

\bibitem{pizer1987adaptive}
Stephen~M Pizer, E~Philip Amburn, John~D Austin, Robert Cromartie, Ari
  Geselowitz, Trey Greer, Bart ter Haar~Romeny, John~B Zimmerman, Karel
  Zuiderveld, {\it Computer vision, graphics, and image processing}
  \textbf{1987}, {\it 39} (3), 355--368.

\end{thebibliography}

% \subsection*{Graphical Abstract Text}
% [Enclose a short descriptive and popular text on the general aim and value of your paper which may serve as an 'appetizer' for the readers (up to 70 words, not a figure caption, not the abstract text). The entry should be written in the present tense and impersonal style. If none is provided, your abstract will be used but truncated for space.]

% \subsection*{Graphical Abstract Figure}
% [Please include a suggestion for an image (full color, size: ideally 5 x 5 cm). It can be one from your article or may be specifically designed for the purpose, such as a photo, schema or drawing. You must hold the copyright to this figure and it should not be reproduced from an external source without permission.]

%DRIVE fig

%STARE fig

%CHASE fig

%HRF fig

%Our fig

\end{document}